\documentclass{article}

\usepackage[toc,page]{appendix}
\usepackage{adjustbox}
\usepackage{wrapfig}
\usepackage{amsmath}
\usepackage{amsfonts}
\usepackage{graphicx}
\usepackage{xcolor}
\usepackage{subcaption}
\usepackage{caption}
\usepackage{xspace}
\usepackage{algorithm}
\usepackage{algorithmic}
\usepackage{listings}
\usepackage{etoc}
\usepackage[preprint]{corl_2026} % Use this for the initial submission.

\newcommand{\cell}[2]{#1 {\scriptsize\textcolor{gray}{(#2)}}}

\newcommand{\para}[1]{\noindent\textbf{#1}}

\newcommand{\ours}{{{Lookahead}}}
\newcommand{\ourengine}{Armory\xspace}

\usepackage{tabularray}
\UseTblrLibrary{booktabs}
\usepackage[font=small,skip=2pt]{caption}

\title{Action Chunk Scheduling for \\ Batched Robot Policy Serving}

\author{
  Rohan Bansal*, David He*, Nadun Ranawaka Arachchige, Zhenyang Chen, Soobum Kim,\\
  \textbf{Kexin Rong$^\dagger$, Danfei Xu$^\dagger$} \\
  * Equal Contribution, $\dagger$ Equal Advising\\
  Georgia Institute of Technology
}

\begin{document}
\maketitle

%===============================================================================

\begin{abstract}
Deploying robot foundation models at scale is the next step towards realizing the potential of general-purpose robots. However, Vision-Language-Action (VLA) and other foundation models are computationally demanding, and on-device compute is constrained by power and space. In this paper, we introduce the problem of serving a robot policy to multiple robots from a remote GPU and formulate it as a scheduling problem. We build \ourengine{}, a serving system validated on fleets of both simulated and real robots. Our experiments show that naive scheduling heuristics perform well when all robots are the same, but fall short when robots consume action chunks at different rates, uncovering a mismatch between conventional batching methods and the closed-loop requirements of robot policy execution. To address this, we propose a scheduling algorithm that accounts for this heterogeneity and improves overall system throughput by up to $18\%$ in real-world experiments. 
Additional details are available at \href{https://gatech-rl2.github.io/actionchunkscheduling/}{https://gatech-rl2.github.io/actionchunkscheduling/}.

 % We present \ourengine{}, a multi-tenant serving engine with a simulation-based scheduler than can effectively navigate these tradeoffs, improving overall system throughput by up to $18\%$ in real-world experiments. We enable a tradeoff that, in some heterogeneous settings, can boost dynamic-task robot throughput by up to $4\times$ without degrading throughput on static-task robots. Additional details are available at \href{https://rohan-bansal.github.io/action-chunk-scheduling/}{this https URL}.

\end{abstract}

%===============================================================================
\vspace{-0.5em}
\section{Introduction}
\vspace{-0.5em}
\label{sec:introduction}
Robot policies are becoming increasingly capable \citep{zitkovich2023rt2,kim2024openvla, black2024pi_0,intelligence2025pi_05,bjorck2025gr00t}, and deploying them at scale turns policy inference into a \emph{policy serving problem}. 
% Recent vision-language-action (VLA) models demonstrate broad manipulation capabilities across tasks, objects, and embodiments . 
In contrast to other machine learning serving problems, robots have strict latency requirements: robots act in closed loop with the physical world and must react quickly to changes in the environment. While local inference avoids network delay, it can be difficult to run foundation-scale policies at real-time rates on power-constrained hardware \citep{jiang2026vlaperf}. Cloud serving is therefore a natural alternative: a powerful GPU can host larger models while amortizing overhead across many robots.

In this paper, we study the multi-tenancy problem for robot policy serving: \emph{how should a single powerful GPU serve many robots through batched inference?} This setting is enabled by action chunking, where each policy query produces a sequence of future actions rather than a single action \citep{zhao2023act}. Since a robot can execute the current chunk while the next query is being processed, action chunking reduces the required inference frequency and makes remote serving practical. Recent  works have proposed \textit{asynchronous execution}, where inference overlaps with execution so that robots can continue execution while waiting for the next chunk \citep{black2025rtc,tang2025vlash,black2025trainrtc}. However, these systems primarily address single-robot serving, where the GPU may remain underutilized between requests. 

\begin{figure}[h]
    \centering
    \includegraphics[width=1\linewidth]{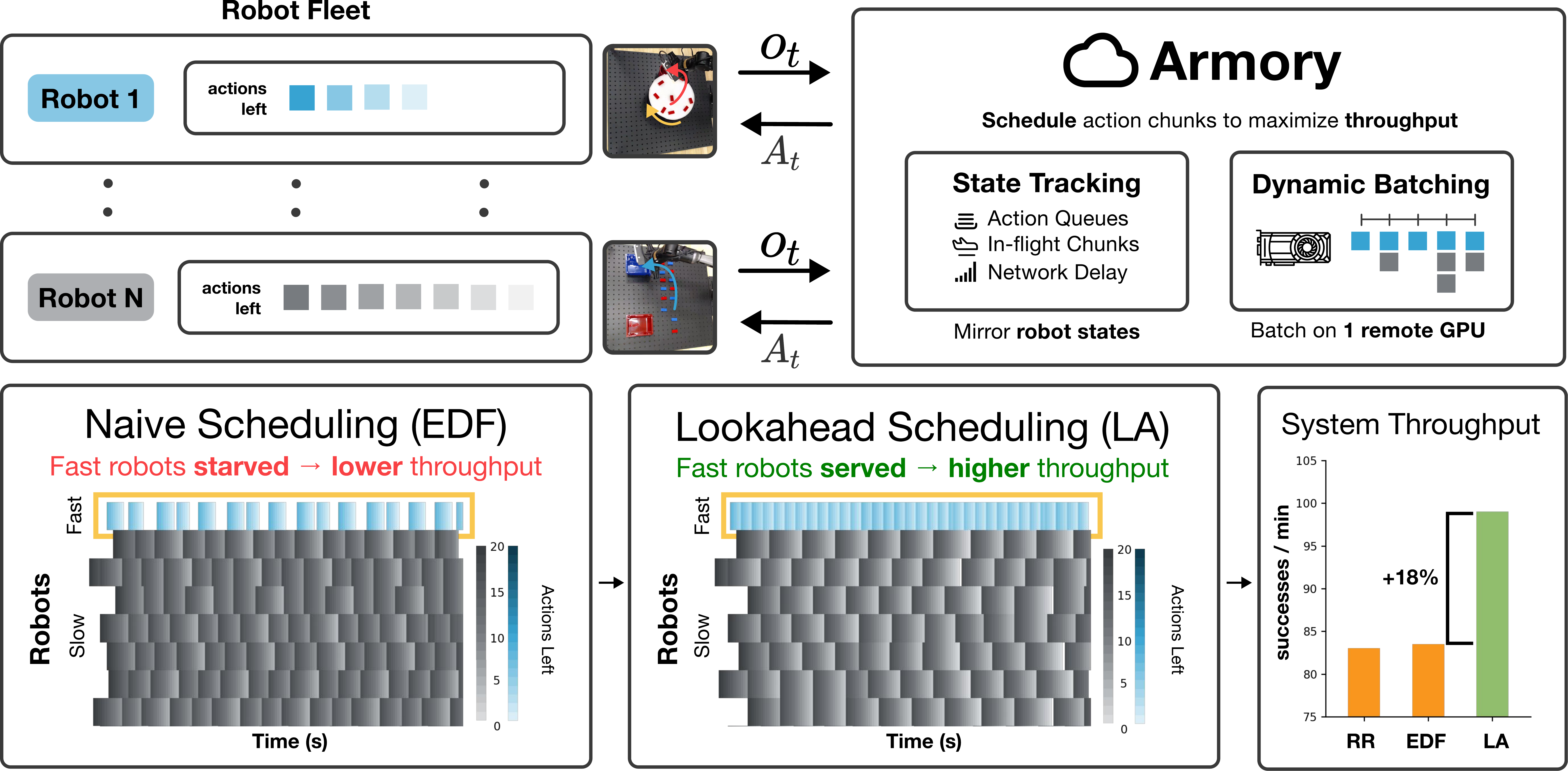}
    \vspace{-2mm}
    \caption{
    % We present batched robot policy serving as a closed-loop serving problem, instantiated as our engine \ourengine{}.
    \ourengine{} is an end-to-end serving system for deploying robot foundation model policies from the cloud that tracks robot states and schedules action chunks to minimize robot starvation. Naive scheduling underserves fast robots completing highly dynamic tasks, so we propose \ours{} scheduling to improve fast robot service with minimal impact on other robots, significantly boosting system throughput under heterogeneous workloads.}
    \label{fig:placeholder}
    \vspace{-1.5em}
\end{figure}

The core difficulty is that batching improves GPU efficiency at the cost of robot responsiveness. Larger batches amortize inference costs, but they also increase latency. Unlike in LLM serving \citep{agrawal2023sarathi}, latency for robotics is not merely a service-level metric. It changes the closed-loop behavior of the robot. If a robot executes a chunk for too long, its actions become stale and it loses reactivity. If the next chunk arrives too late, the robot starves, leaving it with no valid action to execute. This tradeoff is especially difficult for heterogeneous fleets: dynamic tasks may require short execution horizons and frequent updates, while quasi-static tasks may tolerate longer horizons. A system that treats all robots uniformly can therefore underserve the robots whose tasks are most sensitive to delay.

Our key insight is that batched robot policy serving should be formulated as a \emph{closed-loop scheduling problem}.
% , instead of a passive batching problem. 
The server should decide which robots to serve by planning how each batch affects future execution: which robots will continue acting, which will starve.
% , and which need fresher actions because their tasks are more reactive. 
We instantiate this idea in \textbf{\ourengine}, a serving system for batched robot policy inference. \ourengine maintains a server-side mirror of each robot's action queue, in-flight chunks, execution horizon, and communication delay. We formulate action chunk scheduling as a Markov Decision Process in which the server action is a batch of robots to serve and the reward is total executed robot time. Since exact planning is intractable, we propose \ours{}, a practical lookahead scheduler that simulates candidate batches forward and selects the batch with the best predicted reward normalized by inference time.

We evaluate \ourengine in both simulation and on a fleet of 10 real robots. Our experiments show that acceptable execution horizons are task-dependent: dynamic manipulation degrades sharply with stale chunks, while quasi-static manipulation is more tolerant. We also show that starvation is not merely a systems metric; it directly reduces downstream throughput, especially for dynamic tasks. Finally, we find that simple schedulers are competitive in homogeneous fleets, but heterogeneity-aware scheduling is needed when robots have different reactivity requirements. In real-world experiments, lookahead scheduling improves overall system throughput by up to $18\%$ in heterogeneous settings while exposing a controllable tradeoff between serving fast and slow robots.
\vspace{-0.5em}
\section{Preliminaries}
\vspace{-0.5em}
\label{sec:preliminaries}

\para{Action Chunking.} A robot policy processes an input observation $o_t$ to produce an action chunk $A_t = [a_i,a_{i+H-1}]$, where $H$ is the prediction horizon and $i$ is the \textbf{action index} of the first action. During execution, a robot steps at a pre-defined control frequency $f_c$, capturing $o_t$ and executing $a_t$. In \textit{synchronous execution}, a robot executes a chunk to the end of its horizon before generating a new chunk. Generating a new chunk with the policy incurs non-zero delay, creating gaps between executed chunks that appear as pauses or discontinuous motion. To address this, many works adopt \textit{asynchronous execution} \cite{black2025rtc,tang2025vlash}, where the inference for the next chunk starts while the current chunk is still executing. To ensure continuity, executed action indexes must strictly increment \cite{vial2026drtc}.

\para{Execution Horizon.} In practice, a robot only executes a prefix of the full prediction horizon $H$. The length of this prefix should be restricted to an interval $[H_{\text{min}}, H_{\text{max}}]$. If $H_{\text{min}}$ is too small, frequent switching between chunks increases compounding errors~\cite{chi2025diffusion, zhang2025actionchunking}, and if it is too large, robots react too slowly to changes in the environment. Empirically, we find the optimal interval is task-dependent (Figure~\ref{fig:throughput_horizon}), motivating the need to serve workloads with heterogeneous execution horizons.

\begin{figure}[t]
    \centering
    \includegraphics[width=\linewidth]{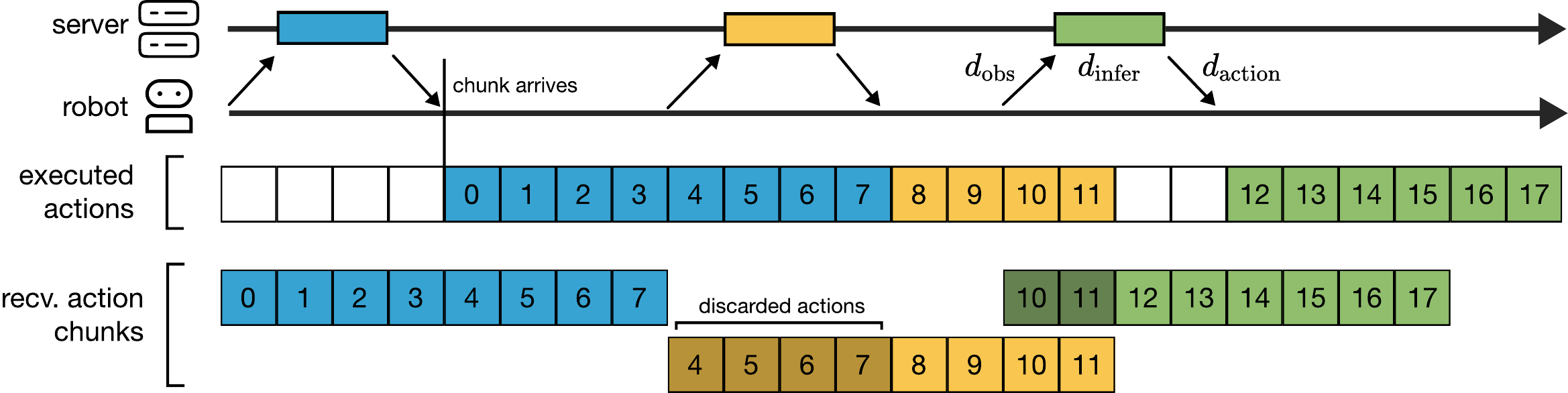}
    \vspace*{1mm}
    \caption{\textbf{Execution Timeline.} A robot with $H_\text{max} = 8$ consumes actions by incrementing the action index, pausing when it is starved. When starvation occurs while a chunk is still in inference/network, the chunk's execution will extend more than $H_\text{max}$ steps after the observation was sent.}
    \label{fig:action_chunks}
    \vspace{-1.3em}
\end{figure}

\para{Delay.} In the real-world, there exists some delay $d$ from the time the robot captures observation $o_t$ to the time it executes an action from the corresponding chunk $A_t$. 
% $d$ can originate from countless sources ranging from sensor latency to system-level jitter. 
We decompose $d$ into three main components: the observation to server delay $d_{\text{obs}}$, the policy inference delay $d_{\text{infer}}$, and the server to action rollout delay $d_{\text{action}}$. 
In cloud serving, $d_{\text{obs}}$ and $d_{\text{action}}$ are dominated by network round-trip time between the robot and the policy server; in on-device or ideal local setups they
are effectively zero.
$d_{\text{infer}}$ is non-negligible in either setting and typically spans multiple control steps.
%Regardless, $d_\text{infer}$ typically covers multiple control steps, which means that starvations are inevitable unless action chunks are created and executed in an asynchronous manner. 

\para{Starvation.} The primary objective of robot policy serving is minimizing the number of control steps where a robot has no actions to execute, which we refer to as a \textit{starvation}. Starvations are harmful because they slow robot execution, induce jerky motions, and impair policy performance. 
As with the execution horizon, we hypothesize the cost of starvation to be task-dependent and validate this empirically in Figure~\ref{fig:throughput_starvation}. 
These task-dependent responses create heterogeneity in inference demands when serving different tasks concurrently, posing challenges we address in Section~\ref{sec:method}.
% \dhe{Starvations are also interesting because they increase the length of action chunks, but how to discuss this nuance?}

\para{Batching.}
%On a GPU, batching the inference of multiple requests 
% is extremely important for effectively using the hardware resources. The latency of inference typically scales sublinearly with batch size because the overhead of loading the model weights is amortized over the computation for many requests. Consequently, larger batch sizes have much higher infer
%significantly improves the inference throughput, measured in requests per second. 
Batching on a GPU significantly increases inference throughput, measured in requests per second, since the cost of loading model weights is amortized across the batch. However,  increasing the batch size increases $d_{\text{infer}}$, delaying the arrival of every generated chunk. 
% For example, increasing the batch size from 1 to 3 for $\pi_{0.5}$ on an L40S increases the throughput from \~14 to \~21 requests/second, but also increases each chunk's latency from 70ms to 140ms.
A good serving system must use \textit{dynamic batching} to carefully balance the tradeoff between throughput and latency.

\vspace{-0.5em}
\section{Scheduling for Batched Robot Policy Serving}
\vspace{-0.5em}
\label{sec:method}

\ourengine{} treats batched policy serving as a control-aware scheduling problem. Action chunk execution depends on when chunks are generated by the server and when they are received by the robot. In the following section, we describe the complex dynamics of action chunk generation and execution, which \ourengine{} models with a server-side software mirror.

% . In the following section,  
% The server does not only schedule inference requests; it schedules action chunks whose value depends on when they are generated and when the robot executes them.
% how they will be used when the robot executes them. 
% Since each robot executes asynchronously while new chunks are being inferred, a scheduling decision changes both the GPU workload and the future action-buffer state of the fleet. 
% \ourengine{} captures this coupling with a server-side execution mirror and uses it to predict the utility of potential batches.
% affect action-buffer continuity, staleness, and starvation.

\vspace{-0.5em}
\subsection{Batched Policy Serving as an MDP}
\label{subsec:mdp}
\vspace{-0.5em}

We formulate batched robot policy serving as a Markov Decision Process. Robots are indexed by $j \in {1,\dots,N}$, and scheduling epochs are indexed by $k$. At epoch $k$, the server observes state $s_k$ and selects a non-empty batch $B_k \subseteq {1,\dots,N}$ of robots to serve. 
% Between epochs, each robot executes actions at control frequency $f_c$.

\para{Server-side execution state.}
For each robot $j$, \ourengine{} tracks three quantities:
\begin{align}
i_j        &\quad \text{action index of the latest executed robot step} \\
\hat{i}_j  &\quad \text{action index of the latest robot step received by the server} \\
\mathcal{Q}_j &\quad \text{queue of generated action chunks} 
% \mathcal{I}_j &\quad \text{in-flight policy requests.}
\end{align}

Inside $\mathcal{Q}_j$, an action chunk for robot $j$ is written as
\begin{equation}
c = (i_{\text{start}},h_j,t_{\text{arr}}),
\end{equation}
where $i_{\text{start}}$ is the action index corresponding to the first action in the chunk, $h_j$ is the robot's execution horizon $H_\text{max}$, and $t_{\text{arr}}$ is the time the chunk arrives at the robot. $\mathcal{Q}_j$ stores all chunks for a robot, including chunks that have been generated by the server but have not yet arrived at the robot.

% Robot $j$ can execute from chunk $c$ at time $\tau$ only if the chunk has arrived and covers the current action index:
% \begin{equation}
% t_{\text{arr}} \leq \tau
% \quad \text{and} \quad
% i_j \in [i_c,; i_c+h_c)
% \end{equation}
% If no chunk in $\mathcal{Q}_j$ satisfies this condition at a control tick, robot $j$ starves.

The full serving state is
\begin{equation}
s_k =
\big(t_k,i_j,\hat{i}_j,\mathcal{Q}_j\big),
% \mathcal{I}_j}*{j=1}^{N}\big),
\end{equation}
where $t_k$ is the wall-clock time at the start of the scheduling epoch. 

% This state is a compact mirror of robot execution, rather than a generic request queue. It records which action indices have already been executed, which observations can still seed useful chunks, which chunks are available, and which requests are already in flight.

\para{Transition.} Let per-batch-size inference latencies be $\tilde{d}_{{\text{infer}}(b)}$ for batch size $b$, and stochastic per-robot networking latencies $\tilde{d}_{\text{obs},j}, \tilde{d}_{\text{action},j}$. Given state $s_k$ and batch $B_k$, inference completes at
\begin{equation}
t_{k+1} = t_k + \tilde{d}_{{\text{infer}}(|B_k|)}.
\end{equation}
% where $\tilde{d}_{{\text{infer}}(|B_k|)}$ is the inference time for the selected batch.
For each served robot $j \in B_k$, the resulting action chunk added to $\mathcal{Q}_j$ is
\begin{equation}
c_{\text{new}}^j =
\big(
\hat{i}_j,
h_j,
t_{k+1}+\tilde{d}_{\text{action},j}
\big).
\end{equation}

During the same interval, all robots tick forward at $f_c$. For each control tick $\tau \in [t_k,t_{k+1}]$, robot $j$ advances by one action index if there is a chunk in $\mathcal{Q}_j$ that has arrived and covers the current index:
\begin{equation}
i_j \leftarrow i_j + 1
\quad \text{if } \exists c \in \mathcal{Q}_j :
t_{\text{arr}} \leq \tau
\wedge
i_j \in [i_c, i_c+h_c).
\end{equation}
% Otherwise, robot $j$ starves at $\tau$. 
The server also updates $\hat{i}_j$ with the latest control step that has arrived at the server by $t_{k+1}$:
\begin{equation}
\hat{i}_j \leftarrow
\max { i : t_i+\tilde{d}_{\text{obs},j} \leq t_{k+1} },
\end{equation}
where $t_i$ is the wall-clock time when action index $i$ was reached on the robot. 
% All delay terms, including $\tilde{d}_{\text{infer}}(B_k)$, $\tilde{d}_{\text{obs},j}$, and $\tilde{d}_{\text{action},j}$, are stochastic.

\para{Reward.}
Let $\Delta i_j(s_k,B_k)$ denote the number of non-starved control steps robot $j$ executes during the transition induced by batch $B_k$.
% This quantity can be computed by rolling forward the server-side execution mirror. 
The per-epoch reward is the total executed robot time:
\begin{equation}
R(s_k,B_k)
=
\frac{1}{f_c}
\sum_{j=1}^{N}
w_j \Delta i_j(s_k,B_k).
\end{equation}
where $w_j$ is an optional task-dependent weight. With $w_j=1$ for all robots, the objective treats all executed robot time equally. Larger $w_j$ values prioritize robots whose tasks require fresher actions, such as dynamic tasks with shorter execution horizons or higher starvation sensitivity.

A scheduling policy $\pi$ maps serving states to batches. The objective is
\begin{equation}
\pi^\star =
\arg\max_\pi
\sum_k
\mathbb{E}\big[ R(s_k,B_k) \big].
\end{equation}

\vspace{-0.5em}
\subsection{Scheduling}
\vspace{-0.5em}
\label{subsec:scheduling}

Exact planning in this MDP is intractable because future batch choices create a branching tree over robot execution states and networking latencies are stochastic. Thus, we propose \textbf{\ours{}}, a scheduling algorithm that plans a few steps ahead using the server-side software mirror. 

At each scheduling epoch, \ours{} enumerates all potential schedules of $L$ epochs. A schedule $S$ is a sequence of states and actions:
\begin{equation}
S = (s_0,B_0),\ (s_1,B_1),\ \dots,\ (s_{L-1},B_{L-1}).
\end{equation}

For each schedule, it rolls the simulation forward and calculates a score $\text{Score}(S)$. \ours{} then dispatches the first batch $B_0$ from the highest scoring schedule. In this work, we consider a score that normalizes the cumulative reward by the time spent on the GPU: 
\begin{equation}
\text{Score}(S)
=
\sum_{\ell=0}^{L-1}
\frac{
R(s_\ell,B_\ell)
}{
d_{\text{infer}}(|B_\ell|)
}.
\end{equation}

We compare Lookahead against two standard scheduling primitives. \textbf{Round Robin (RR)} cycles through robots in batches of maximum size $b$, serving robots fairly but ignoring urgency. \textbf{Earliest Deadline First (EDF)} schedules the $b$ robots predicted to run out of executable actions soonest. EDF captures imminent starvation, but treats all starvation events as equally costly and does not model how the chosen batch affects future serving states. Lookahead instead evaluates batches by their induced execution trajectory, allowing the scheduler to trade off GPU efficiency, starvation avoidance, and task-dependent reactivity.

\vspace{-0.5em}
\subsection{Armory}
\vspace{-0.5em}
\label{subsec:armory}

We implement \ourengine{}, an end-to-end serving engine for batched robot policy inference, designed around the software mirror of the MDP described in \ref{subsec:mdp}. To prioritize mirror accuracy, \ourengine{} uses a push-based communication architecture, where each robot sends the captured observation $o_t$ before executing action $a_t$. The server tracks the latest received action index $\hat{i}_j$, queued chunks $\mathcal{Q}_j$. 
Latencies $d_{\text{infer}}$ are profiled on startup and $d_{\text{obs}}$ and $d_{\text{action}}$ are continuously estimated for each robot.
% Incoming observations are organized per robot rather than as a single global FIFO queue. At each scheduling epoch, the scheduler selects robots to serve. For each selected robot, the engine uses the most recent useful observation and drops older observations that would generate chunks for stale action indices. This behavior is intentional: in asynchronous robot serving, processing every observation is less important than producing chunks that can still arrive in time to affect execution.

% \ourengine{} separates network handling, scheduling, and GPU inference into independent processes so that CPU-side networking and scheduling do not block GPU execution. Network workers receive observations and return completed chunks. The scheduler process maintains the execution mirror and commits batches. The GPU worker runs policy forward passes. The scheduler is work-conserving: whenever the GPU is available and at least one useful request exists, it dispatches a batch. Additional implementation details are in Appendix \ref{sec:implementation}.

\ourengine{} separates network handling, scheduling, and GPU inference into independent processes communicating through shared memory, so that CPU-side work never blocks the GPU. The scheduler maintains the execution mirror and is work-conserving: whenever the GPU is idle and at least one useful request exists, it dispatches a batch. More implementation details are in Appendix~\ref{sec:implementation}.
\vspace{-0.5em}
\section{Experiments}
\vspace{-0.5em}

\begin{wrapfigure}{r}{0.3\textwidth}
    \centering

    \begin{subfigure}{0.95\linewidth}
        \centering
        \includegraphics[width=\linewidth]{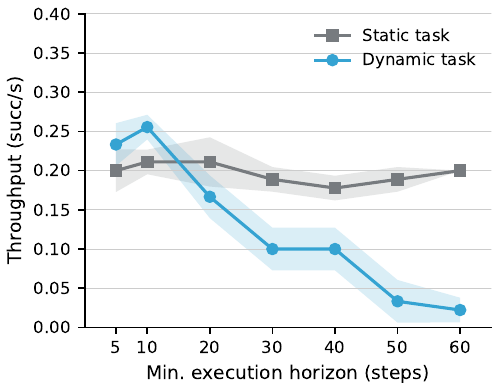}
        \caption{Execution horizon.}
        \label{fig:throughput_horizon}
    \end{subfigure}

    \vspace{0.4em}

    \begin{subfigure}{0.95\linewidth}
        \centering
        \includegraphics[width=\linewidth]{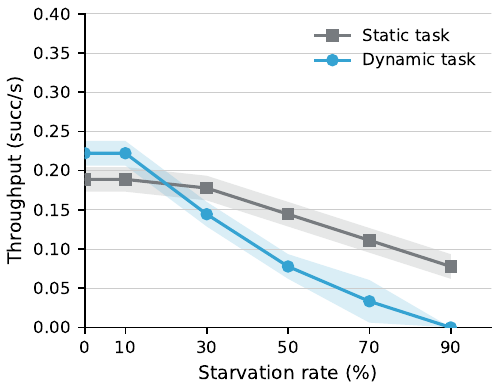}
        \caption{Starvation rate.}
        \label{fig:throughput_starvation}
    \end{subfigure}

    \caption{Dynamic task throughput is more sensitive to stale execution horizons and starvation than that of quasi-static tasks.}
    \label{fig:throughput_effects}
    \vspace{-2em}
\end{wrapfigure}

In this section, we describe the evaluation of our suite of scheduling algorithms in real-world and simulation experiments under different levels of robot heterogeneity.
We find that under a homogeneous workload, simpler algorithms may suffice, while under heterogeneous workloads, there is a tradeoff between overall system starvation and throughput. For all evaluations, we use $\pi_{0.5}$ \cite{intelligence2025pi_05} as a representative robotics foundation model.

\vspace{-0.5em}
\subsection{Setup}
\vspace{-0.5em}

% \begin{wrapfigure}{r}{0.33\textwidth}
%   \centering
%   \begin{tabular}{ccc}
%     \toprule
%     \textbf{Batch Size} & \textbf{$d_\text{infer}$ (ms)} \\
%     \midrule
%     1 & 73.0 \\
%     2 & 109.8 \\
%     3 & 142.4 \\
%     4 & 177.5 \\
%     5 & 211.1 \\
%     \bottomrule
%   \end{tabular}
%   \caption{L40S inference latency.}
%   \label{tab:l40s_latency}
%   \vspace{-1em}
% \end{wrapfigure}

% \begin{figure}[h]
%     \centering

%     \begin{subfigure}[t]{0.3\textwidth}
%         \centering
%         \includegraphics[width=\linewidth]{figures/throughput_vs_horizon.pdf}
%         \caption{Execution horizon.}
%         \label{fig:execution_horizon}
%     \end{subfigure}
%     \hspace{2em}
%     \begin{subfigure}[t]{0.3\textwidth}
%         \centering
%         \includegraphics[width=\linewidth]{figures/throughput_vs_starvation.pdf}
%         \caption{Starvation rate.}
%         \label{fig:starvation}
%     \end{subfigure}

%     \caption{We finetune $\pi_{0.5}$ on static and dynamic pick-and-place tasks. 
%     (a) As the execution horizon increases, policy performance degrades on the dynamic task but remains stable for the static task. 
%     (b) Throughput drops faster under starvation in dynamic tasks than in quasi-static settings because starvation both delays execution and reduces reactivity.}
%     \label{fig:throughput_effects}
% \end{figure}

\para{Workload and Heterogeneity.} Not all robots are equal: some must be more reactive than others. We capture this through two robot classes that differ in execution horizon. A \textit{fast} robot runs what we call a "dynamic" task with a shorter $H_{\text{max}}$, since policy performance degrades quickly when chunks are stale while manipulating moving objects; a \textit{slow} robot tolerates longer horizons for quasi-static manipulation (Figure~\ref{fig:throughput_horizon}), running what we call "static" tasks. Dynamic tasks are sensitive to loss of reactivity, so starvation also hurts fast robots disproportionately (Figure~\ref{fig:throughput_starvation}). Note that \textit{fast} and \textit{slow} refer to the speed at which a robot exhausts its chunks as a result of its horizon, and \textbf{not} its movement/step speed; robots that exhaust chunks faster are more reactive for this reason.

We vary heterogeneity across three configurations: \textit{one-fast} (one fast robot, rest slow), \textit{half-fast} (half fast, half slow), and \textit{all-fast} (all robots fast, none slow). These span the heterogeneity spectrum and stress the scheduler in distinct ways. \textit{All-fast} is a homogeneous setting that isolates the effect of fleet size from heterogeneity. \textit{Half-fast} maximizes scheduling tension: fast and slow robots compete equally for batch slots, so bias toward one class visibly starves the other. \textit{One-fast} tests whether the scheduler can protect a single sensitive robot from many less urgent ones; a realistic analog is a robot that may become highly dynamic for a short period (e.g., striking a match) before resuming slower operation.

\para{Server Specifications.} For both simulation and real-world experiments, the cloud server is colocated on the network (round-trip delay $\sim$ 5–10 ms) and uses a single L40S GPU. We show in Appendix \ref{sec:network_ablate} that serving remains (1) practical with network delay up to 50 ms (a typical US coast-to-coast round trip), beyond which added delay simply acts as a stronger starvation signal, and (2) robust to network jitter from mild to extreme.

% \begin{figure}[t]
%     \centering
%     \includegraphics[width=\linewidth]{figures/tier_breakdown_combined_aligned__mbs3.pdf}
%     \vspace*{-2mm}
%     \caption{LIBERO-10 simulation results. For one-fast, LA@5 is able to maintain a consistently higher throughput and lower starvation for fast-tier robots without sacrificing slow-tier performance. For half-fast, LA@3 is the most useful operating point: it improves fast-tier service while keeping the slow-tier close to the fixed-priority baselines. LA@5 shows the boundary of the tradeoff, where further prioritizing fast robots begins to significantly sacrifice slow-tier performance.}
%     \label{fig:sim_results_2_3}
%     \vspace{-1.5em}
% \end{figure}

\begin{figure}[t]
    \centering
    \includegraphics[width=\linewidth]{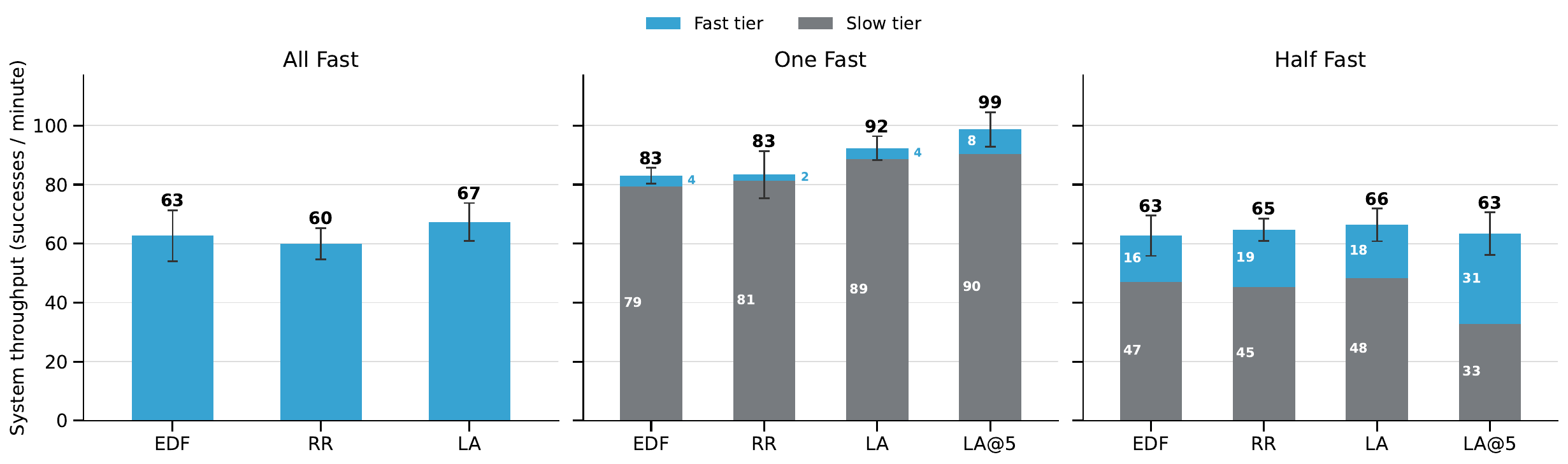}
    \vspace*{-2mm}
    \caption{We compare system throughput in legos-per-minute across the three heterogeneous scenarios in a real-world setting. We observe that LA@5 significantly improves fast-tier throughput in the one-fast case by almost $2\times$ over EDF, and provides a small boost to slow-tier as well, increasing system throughput by approximately $18\%$ over baselines. In the half-fast case, we can maintain system throughput but trade off towards fast robots, boosting their share of system throughput significantly. Counterintuitively, there are some cases when throughput increases despite an increase in starvation, which we analyze in Appendix \ref{sec:real_metrics}.}
    \vspace{-1.7em}
    
    % \kr{add red vs blue bar legend in figure. also consider drawing this as a stacked bar chart, so the total bar height shows system throughput. i was looking for the 18\% throughput improvement number }}
    \label{fig:throughput_real_all}
\end{figure}

\para{Simulation.} Our cloud server runs a $\pi_{0.5}$-LIBERO checkpoint ($H=10$, $f_c=20$) on a node separate from the simulation clients. Both classes set $H_{\text{min}}=1$; \textit{fast} robots use $H_{\text{max}}=6$ and \textit{slow} robots use $H_{\text{max}}=10$. We develop a LIBERO~\cite{liu2023libero} evaluation harness that steps each robot independently in real time. Each robot is assigned one of two randomly sampled LIBERO-10 tasks and completes as many episodes as possible in a 5-minute window (30-second time-out per episode). Heterogeneity is applied per task, where task 1 robots are \textit{fast}, task 2 robots are \textit{slow} (excluding all-fast, where all tasks are fast). We sweep 3 seeds and fleet size from 2 to 10 across all 3 configurations.

\begin{wrapfigure}{r}{0.42\textwidth} % r = right, l = left
    \vspace{-1.4em}
    \centering
    \includegraphics[width=\linewidth]{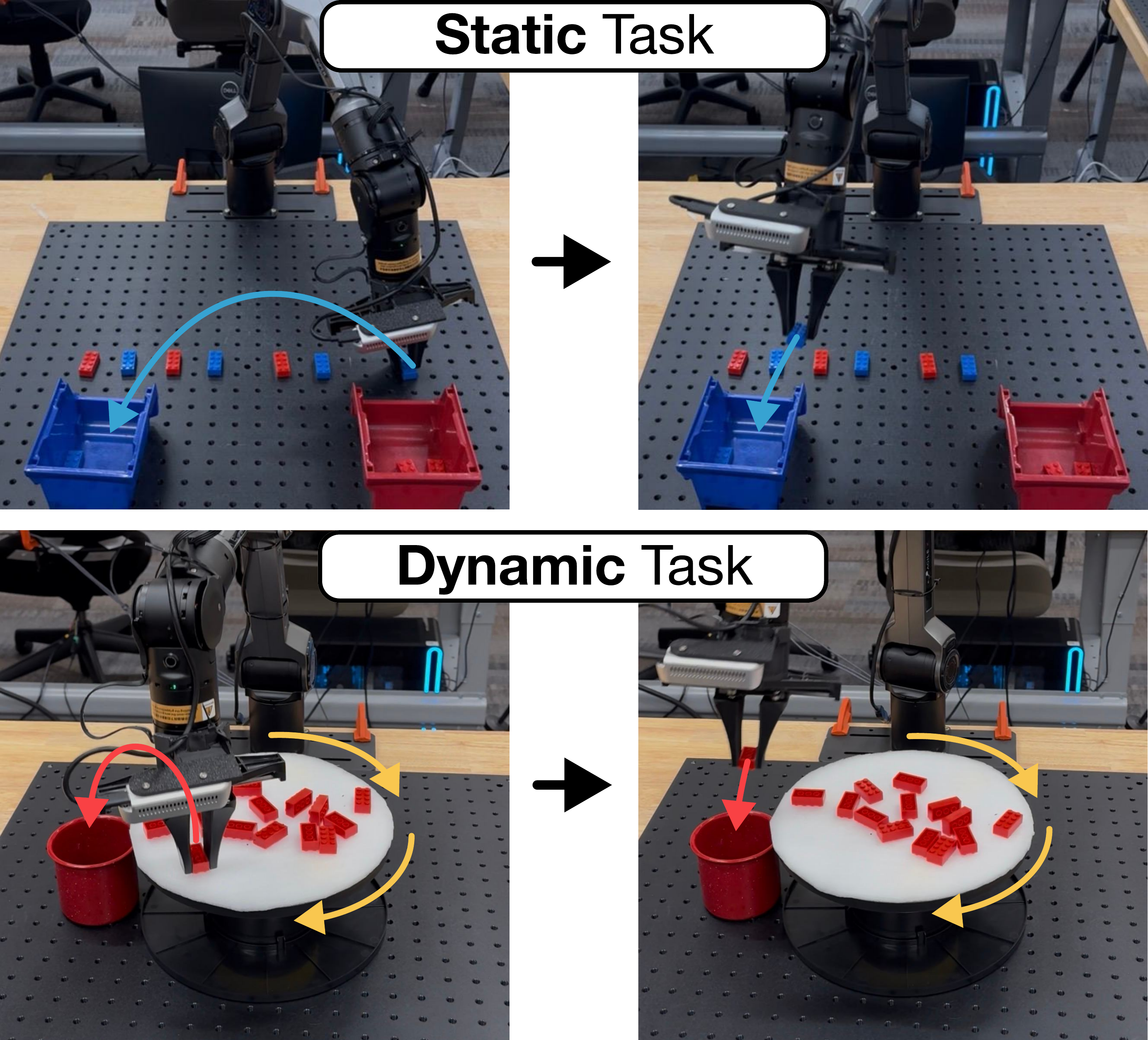}
    \vspace{-0.6em}
    \caption{Real-world task setup. Top is the static brick-into-bin sorting task, and bottom is the dynamic brick-into-mug turntable task.}
    \label{fig:real_world_setup}
    \vspace{-1.5em}
\end{wrapfigure}

%Each robot within a scenario is assigned one of two randomly sampled tasks in LIBERO-10. Each robot will work to complete its task as many times as possible within a 1 minute time limit, given normal per-task constraints (cannot exceed 600 execution steps). Heterogeneity is applied per-task; robots with task 1 will be \textit{fast}, and robots with task 2 will be \textit{slow} (excluding 10 fast, where all tasks are fast).

\para{Real World.} We deploy 10 AgileX PiPER arms with top and wrist cameras (Intel RealSense D435i), served by a $\pi_{0.5}$ checkpoint finetuned at $H=20$ and $f_c=30$. Both classes use $H_{\text{min}}=5$ to prevent excessively multimodal motions; \textit{fast} robots use $H_{\text{max}}=10$ and \textit{slow} robots use $H_{\text{max}}=20$. We fine-tune on two tasks: static brick sorting (sorting colored bricks into bins) and dynamic turntable sorting (picking moving bricks off a turntable into a mug). Fast robots run the turntable task and slow robots run bin sorting, except in the homogeneous setting where all run bin sorting. 
% We use this special homogeneous case so that all robots share the same task and execution-horizon requirements, allowing us to test whether sophisticated scheduling is still useful when there is no task-induced heterogeneity. 
We evaluate 3 seeds per scheduler per configuration, similar to sim.

% \kr{need to be careful with this. the motivation is about serving from the cloud so it's unexpected to see 0 network delay in eval. this raises questions like how did you run the real world experiments / whether serving is actually practical with real-world network delay. }

%On real hardware (AgileX PiPER), we finetune $\pi_{05}$ for an action horizon of 20 running at 30Hz: both classes use a minimum execution horizon of 5, while fast robots use a maximum of 10 and slow robots the full 20. 

%For the homogeneous setting, we compare round-robin, max-batch, and \ours{} across 3 seeds. For the heterogeneous setting, we evaluate 1 fast 9 slow and 5 fast 5 slow to showcase the throughput tradeoff capability of \ours{}. 

%In real, we fine tune $\pi_{05}$ on two tasks, static brick sorting and dynamic turntable brick sorting. In the former, the robot must sort red and blue bricks on the table into the correctly colored bin. In the latter, the robot must pick moving bricks off a turntable and place them in a mug on the table. We use the turntable task for all fast robots, excluding the 10 fast case in which all robots use bin sorting.

\para{Schedulers.} For all experiments, we analyze the Round-Robin, Earliest-Deadline-First, and \ours{} schedulers described in \ref{subsec:scheduling}. In simulation, we set $b=3$, and in real, we set $b=3$ for the homogeneous and $b=5$ for the heterogeneous setting. These choices favor RR and EDF; we sweep batch sizes to find the best-performing ones for these two schedulers to make a fair comparison. See Appendix \ref{sec:batch_ablate} for detailed ablations on batch size. 

To study the tradeoff ability of \ours{}, we set $w_j = 1$ for all slow robots and compare three weightings $w_j=1,3,5$ for the fast robots, each denoted as LA@$w_j$. Intuitively, $w=1$ treats all robots equally, $w=3$ moderately prioritizes robots with shorter horizons and higher starvation sensitivity, and $w=5$ is an aggressive setting for deployments where protecting fast robots is worth sacrificing slow-tier service. Although the formulation permits multi-step receding-horizon planning, we find that a one-step model-based score is sufficient for heterogeneity-aware scheduling (Appendix \ref{sec:lookahead_ablate}), so we use the simplest instantiation of the framework with $L=1$.

% In \ref{sec:lookahead_ablate}, we experimented with deeper rollout horizons but found that $L=1$ was sufficient to achieve the desired heterogeneity-aware scheduling behavior.

\para{Metrics.} We report two primary metrics: \textit{starvation rate}, the fraction of control steps where a robot has no action to execute, and \textit{throughput}, the number of successes a robot produces in a unit time. 
In simulation, throughput is the number of successful task completions each robot achieves per minute of wall-clock time. In the real world, throughput is the number of Lego pieces each robot sorts during a fixed 60-second window (also per-minute).
%In simulation, throughput is calculated by dividing the total number of successful task completions for each robot by the total wall-clock seconds observed, yielding a per-robot successes-per-second rate. 
%In the real-world experiments, throughput is measured directly by counting the number of Lego pieces each robot completes within a fixed 60-second window, yielding a per-robot legos-per-minute rate.

% \subsection{Results}
% \begin{figure}
%     \centering
%     \includegraphics[width=1\linewidth]{figures/average_starvation_simulation.pdf}
%     \caption{Enter Caption}
%     \label{fig:placeholder}
% \end{figure}

\begin{figure}[t]
    \centering
    \includegraphics[width=\linewidth]{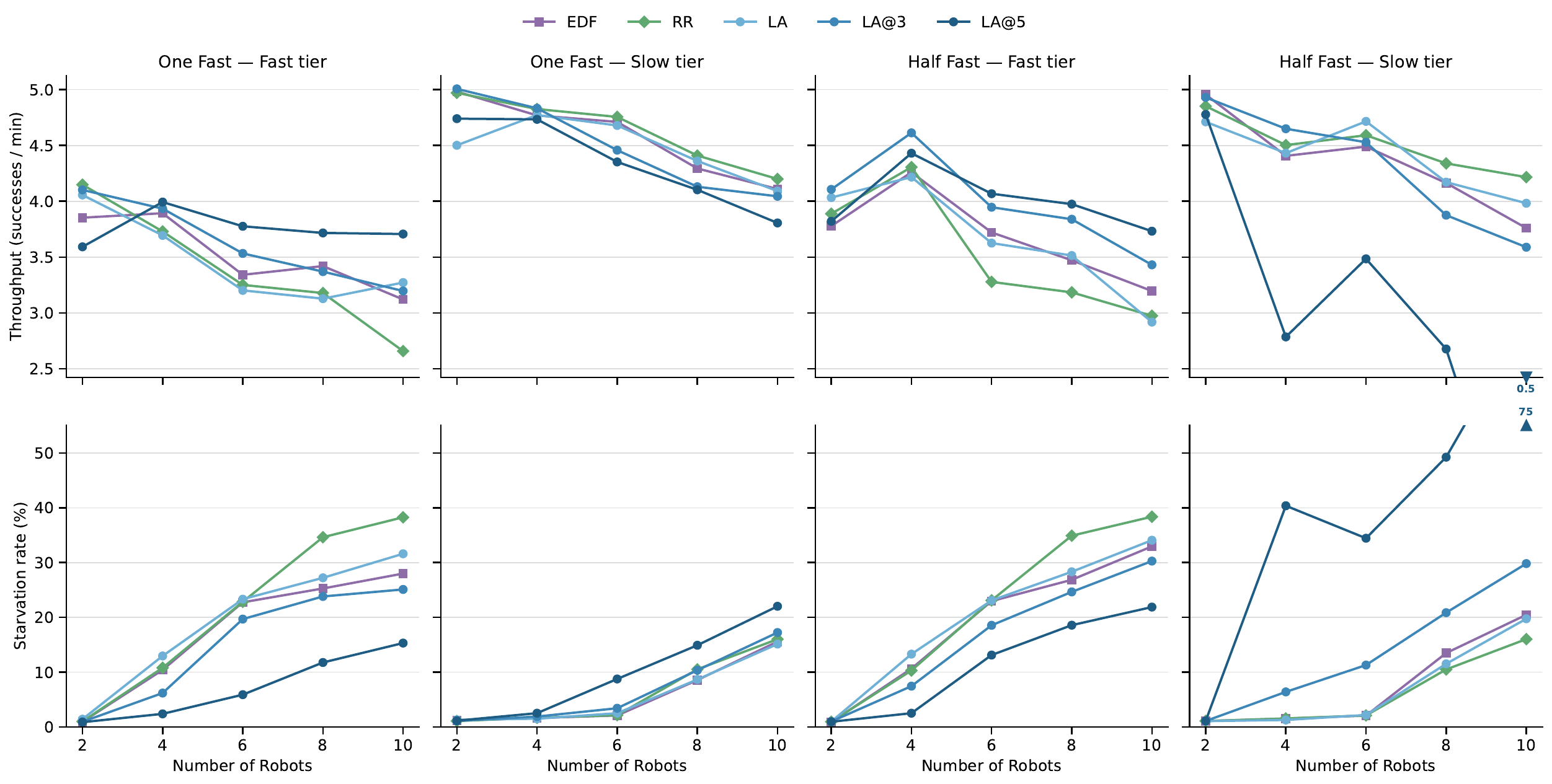}
    \vspace*{-2mm}
    \caption{LIBERO-10 simulation results. For one-fast, LA@5 is able to maintain a consistently higher throughput and lower starvation for fast-tier robots without sacrificing slow-tier performance. For half-fast, LA@3 is the most useful operating point: it improves fast-tier service while keeping the slow-tier close to the fixed-priority baselines. LA@5 shows the boundary of the tradeoff, where further prioritizing fast robots begins to significantly sacrifice slow-tier performance.}
    \label{fig:sim_results_2_3}
    \vspace{-1.5em}
\end{figure}

\vspace{-0.7em}
\subsection{Evaluations}
\label{subsec:evaluations}
\vspace{-0.5em}

\para{Weighted scheduling exposes a controllable throughput tradeoff.} In simulation, we sweep fleet size over the three scenarios. In the homogeneous all-fast setting, all methods perform comparably; we delegate these results to Appendix \ref{sec:sim_metrics} and focus on the more revealing heterogeneous settings.
When fast and slow robots share a GPU, the scheduler must choose between maximizing average service and protecting robots that exhaust chunks faster. Figure~\ref{fig:sim_results_2_3} decomposes throughput and starvation by robot tier. In the one-fast setting, weighting the fast robot higher allows \ours{} to serve it before it starves, improving fast-tier throughput while preserving slow-tier performance thanks to longer action buffers. In the half-fast setting, the same weighting creates a sharper tradeoff: moderate weighting improves fast-tier throughput with limited slow-tier degradation, while aggressive weighting over-serves fast robots at the expense of slow ones. This is desirable for an operator-facing scheduler: the weight parameter controls throughput allocation between robot classes, and in some scenarios (namely one-fast), \textit{we can gain throughput for free}. Appendix \ref{sec:real_metrics} reports per-tier metrics as fleet size increases.

% \para{Weighted scheduling transfers from simulation to real.} In the real world, we observe the throughput gains observed in simulation on a fleet of ten real robots (Figure~\ref{fig:throughput_real_all}) for homogeneous and heterogeneous deployments.
% In the homogeneous all-fast setting, \ours{} achieves the highest measured throughput, but all schedulers are broadly comparable because every robot has the same task and horizon.
% This matches the simulation result: when the workload is homogeneous, simple scheduling rules can be sufficient. 

% In heterogeneous deployments, \ours{} provides a more useful control surface. In the 1-fast setting, \ours{} with a weight of 5 can boost the fast-tier throughput by $4\times$ over the RR baseline, and $2\times$ the EDF baseline, while preserving the throughput of the slow robots well. In the half-fast setting, the same aggressive weighting improves fast-tier throughput but correspondingly decreases slow-tier throughput, showing a more aggressive tradeoff that can be made at the discretion of how prioritized dynamic tasks should be.

% Across both simulation and real hardware, these results show that the main benefit of \ours{} is not that it universally dominates a fixed baseline, but that it uses a model of action-buffer dynamics to choose the desired point on the throughput/starvation tradeoff.

\para{Scheduling gains transfer from simulation to real hardware.} On a fleet of ten PiPER arms (Figure~\ref{fig:throughput_real_all}), the homogeneous setting again shows all schedulers performing comparably. In heterogeneous deployments, \ours{} provides a meaningful control surface: in the one-fast setting, \ours{} at $w=5$ boosts fast-tier throughput by $4\times$ over RR and $2\times$ over EDF while preserving slow-tier throughput. In the half-fast setting, the same weighting improves fast-tier throughput significantly at the cost of slow-tier service, exposing a sharper tradeoff that operators can tune to their priorities. Across both settings, the main benefit of \ours{} is not universal dominance over fixed baselines, but the ability to use a model of action-buffer dynamics to select a desired operating point on the throughput/starvation tradeoff.
\vspace{-0.5em}
\section{Related Work}
\vspace{-0.5em}

\label{sec:related work}

\para{Robot Policy Inference.}
Robot foundation models are rapidly scaling: early VLA systems~\citep{zitkovich2023rt2} showed that large VLM backbones could be adapted to robotic control, and more recent models pair even larger backbones with broader cross-embodiment datasets~\citep{kim2024openvla, black2024pi_0, intelligence2025pi_05, kim2026cosmospolicy, bjorck2025gr00t, o2024open, khazatsky2024droid}. This trend intensifies as the field expands to video prediction, with state-of-the-art models taking up to 7 seconds per inference~\citep{ye2026dreamzero}. On the systems side, ~\citep{jiang2026vlaperf} profiles the bottlenecks of VLA inference,~\citep{kwok25robomonkey} performs inference-time verification, and~\citep{ma2025realtimevla} pushes latency down through kernel-level optimization, while~\citep{hoeg2024streamingdiffusionpolicy} modifies policy formulation to synthesize actions faster. Closer to our setting, several methods improve latency tolerance by overlapping inference with execution through real-time chunking~\citep{black2025rtc, black2025trainrtc}, reasoning over future state during async inference~\citep{tang2025vlash}, accelerating rollout~\citep{guo2025demospeedup}, or combining action scheduling with controller tuning~\citep{arachchige2025sail}. These efforts, however, are orthogonal: they make individual policies faster or more latency-tolerant, whereas we schedule shared compute across multiple robots and are architecture-agnostic, benefiting from any per-model speedup.

\para{ML Serving Systems.}
Early work on low-latency prediction serving and pipeline provisioning established modular serving abstractions, cost-aware configuration, and latency-SLO management for generic inference workloads~\citep{crankshaw2017clipper, crankshaw2018inferline, gujarati2020clockwork}. More recent systems for transformer and LLM serving introduce iteration-level scheduling, efficient memory management, prefix caching, chunked prefills, and disaggregated execution to improve throughput and tail latency~\citep{yu2022orca, kwon2023vllm, zheng2024sglang, agrawal2023sarathi, zhong2024distserve}. While many of these systems may optimize for latency, they do not serve models where the latency directly affects model performance. Robot policy serving is uniquely challenging in that delayed action chunks can directly lead to downstream execution failures.
% These systems primarily optimize text generation or generic inference services; they do not model the consequences of delayed actions, stale observations, or starvation in robot control, which are constraints that our scheduler is designed to operate within.
% \dhe{need to mention that robots continuously consume actions, do not model this deadline, nor do they model }

\para{Real-Time Scheduling.}
Classical real-time scheduling studies how recurring tasks can be ordered to satisfy timing constraints, with foundational formulations dating back to 1973~\citep{liu1973scheduling}. In adjacent systems settings,~\citep{tumanov2016tetrisched} introduces plan-ahead rescheduling for deadline-aware jobs in dynamic heterogeneous clusters, while ~\citep{narayanan2020gavel} studies heterogeneity-aware scheduling for shared accelerator clusters running machine learning workloads. These works provide useful abstractions for deadlines, resource contention, and heterogeneous compute allocation. However, they do not address robot policy serving, where inference requests are coupled to closed-loop control, network, and action-chunk starvation; our formulation adapts the receding-horizon idea to this setting, where the cost of a scheduling decision depends on the physical state of each robot.

Two recent works also target multi-robot serving. Kairos~\citep{dai2026kairos} minimizes end-to-end task latency through variable execution horizon and dynamically prioritizing action generation and execution. ROSA~\citep{jiang2026rosa} maximizes action throughput in robot factories through a static provisioning schedule for known workloads, and further assumes synchronous execution. Both treat latency as a target, and neither models the closed-loop consequence of a late chunk, i.e. the downstream starvation and execution cost it induces. This gap matters heavily under heterogeneity, the differing rates at which robots exhaust chunks and starve: our work folds starvation and closed-loop control into an MDP that allows us to schedule against each robot's execution state and protect different classes of robots.

\vspace{-0.5em}
\section{Conclusion}
\vspace{-0.5em}
\label{sec:conclusion}

In this work, we introduce multi-tenant robot policy serving as a closed-loop scheduling problem. By formulating the server's batching decisions as an MDP over robot action queues, execution horizons, and communication delays, we showed that policy serving has structure that generic batching strategies leave on the table. Our experiments in simulation and on a fleet of ten real robots confirm three findings. First, the cost of latency in robot serving is task-dependent: dynamic manipulation degrades sharply with stale chunks and starvation, while quasi-static tasks are more tolerant. Second, when all robots share the same timing requirements, simple schedulers like Round Robin and Earliest Deadline First perform comparably to more sophisticated alternatives. Third, when robots have heterogeneous reactivity demands, a \ours{} scheduler that simulates action chunking dynamics can improve system throughput by up to 18\% in the real world by selectively protecting latency-sensitive robots, with tunable weights that let operators choose which robots to prioritize.

\vspace{-0.5em}
\section{Limitations}
\vspace{-0.5em}
In recent years, many works have proposed inference optimizations for large transformer models. Notable techniques include continuous batching~\cite{yu2022orca}, PagedAttention~\cite{kwon2023vllm}, chunked prefill~\cite{agrawal2023sarathi}, and prefill-decode disaggregation~\cite{zhong2024distserve, patel2024splitwise}. State-of-the-art serving systems also typically implement distributed inference to support larger models. In this work, we assumed the simplest inference model (batching on a single GPU). We hypothesize there is more room for improvement by co-designing action chunk scheduling together with these inference optimizations.

% In this work, we also assume a very simple execution strategy for the robot. For one, the execution horizon interval $[H_{\text{min}}, H_{\text{max}}]$ is varied across robots, but fixed for a robot's lifetime. 
% However, robots are likely to require varying amounts of reactivity over their lifetime, so future works may investigate varying these parameters.

\acknowledgments{The real-world experiments presented in this work were conducted in the Advanced Robotic Manipulation (ARM) Lab at Georgia Tech.}

\clearpage
\newpage
\clearpage
% The acknowledgments are automatically included only in the final and preprint versions of the paper.

%===============================================================================

\bibliography{example}  % .bib

@article{liu1973scheduling,
  title={Scheduling algorithms for multiprogramming in a hard-real-time environment},
  author={Liu, Chung Laung and Layland, James W},
  journal={Journal of the ACM (JACM)},
  volume={20},
  number={1},
  pages={46--61},
  year={1973},
  publisher={ACM New York, NY, USA}
}

@inproceedings{tumanov2016tetrisched,
  title={{TetriSched}: Global rescheduling with adaptive plan-ahead in dynamic heterogeneous clusters},
  author={Tumanov, Alexey and Zhu, Timothy and Park, Jun Woo and Kozuch, Michael A and Harchol-Balter, Mor and Ganger, Gregory R},
  booktitle={Proceedings of the 11th European Conference on Computer Systems},
  year={2016},
  publisher={Association for Computing Machinery}
}

@inproceedings{narayanan2020gavel,
  title={Heterogeneity-aware cluster scheduling policies for deep learning workloads},
  author={Narayanan, Deepak and Rao, Ratul and Kandula, Srikanth and Seshadri, Aakanksha and Roberts, Sasha and Chaudhary, Parul and Gu, Jason and Gonzalez, Joseph and Harlap, Aaron and Krishnamurthy, Arvind and others},
  booktitle={14th USENIX Symposium on Operating Systems Design and Implementation (OSDI 20)},
  pages={481--498},
  year={2020},
  publisher={USENIX Association}
}

@inproceedings{crankshaw2017clipper,
  title={Clipper: A Low-Latency Online Prediction Serving System},
  author={Crankshaw, Daniel and Wang, Xin and Zhou, Giulio and Franklin, Michael J and Gonzalez, Joseph E and Stoica, Ion},
  booktitle={14th USENIX Symposium on Networked Systems Design and Implementation (NSDI 17)},
  year={2017},
  publisher={USENIX Association}
}

@article{crankshaw2018inferline,
  title={InferLine: ML Prediction Pipeline Provisioning and Management for Tight Latency Objectives},
  author={Crankshaw, Daniel and Sela, Gur-Eyal and Zumar, Corey and Mo, Xiangxi and Gonzalez, Joseph E and Stoica, Ion and Tumanov, Alexey},
  journal={arXiv preprint arXiv:1812.01776},
  year={2018}
}

@inproceedings{gujarati2020clockwork,
  title={Serving DNNs Like Clockwork: Performance Predictability from the Bottom Up},
  author={Gujarati, Arpan and Karimi, Reza and Alzayat, Safya and Hao, Wei and Kaufmann, Antoine and Vigfusson, Ymir and Mace, Jonathan},
  booktitle={14th USENIX Symposium on Operating Systems Design and Implementation (OSDI 20)},
  pages={443--462},
  year={2020},
  publisher={USENIX Association}
}

@inproceedings{yu2022orca,
  title={Orca: A distributed serving system for $\{$Transformer-Based$\}$ generative models},
  author={Yu, Gyeong-In and Jeong, Joo Seong and Kim, Geon-Woo and Kim, Soojeong and Chun, Byung-Gon},
  booktitle={16th USENIX symposium on operating systems design and implementation (OSDI 22)},
  pages={521--538},
  year={2022}
}

@inproceedings{kwon2023vllm,
  title={Efficient memory management for large language model serving with pagedattention},
  author={Kwon, Woosuk and Li, Zhuohan and Zhuang, Siyuan and Sheng, Ying and Zheng, Lianmin and Yu, Cody Hao and Gonzalez, Joseph and Zhang, Hao and Stoica, Ion},
  booktitle={Proceedings of the 29th symposium on operating systems principles},
  pages={611--626},
  year={2023}
}

@article{zheng2024sglang,
  title={Sglang: Efficient execution of structured language model programs},
  author={Zheng, Lianmin and Yin, Liangsheng and Xie, Zhiqiang and Sun, Chuyue Livia and Huang, Jeff and Yu, Cody Hao and Cao, Shiyi and Kozyrakis, Christos and Stoica, Ion and Gonzalez, Joseph E and others},
  journal={Advances in neural information processing systems},
  volume={37},
  pages={62557--62583},
  year={2024}
}

@inproceedings{zhong2024distserve,
  title={$\{$DistServe$\}$: Disaggregating prefill and decoding for goodput-optimized large language model serving},
  author={Zhong, Yinmin and Liu, Shengyu and Chen, Junda and Hu, Jianbo and Zhu, Yibo and Liu, Xuanzhe and Jin, Xin and Zhang, Hao},
  booktitle={18th USENIX Symposium on Operating Systems Design and Implementation (OSDI 24)},
  pages={193--210},
  year={2024}
}

@inproceedings{patel2024splitwise,
  title={Splitwise: Efficient generative llm inference using phase splitting},
  author={Patel, Pratyush and Choukse, Esha and Zhang, Chaojie and Shah, Aashaka and Goiri, {\'I}{\~n}igo and Maleki, Saeed and Bianchini, Ricardo},
  booktitle={2024 ACM/IEEE 51st Annual International Symposium on Computer Architecture (ISCA)},
  pages={118--132},
  year={2024},
  organization={IEEE}
}

@article{agrawal2023sarathi,
  title={Sarathi: Efficient llm inference by piggybacking decodes with chunked prefills},
  author={Agrawal, Amey and Panwar, Ashish and Mohan, Jayashree and Kwatra, Nipun and Gulavani, Bhargav S and Ramjee, Ramachandran},
  journal={arXiv preprint arXiv:2308.16369},
  year={2023}
}

@article{jiang2026vlaperf,
  title={How Fast Can I Run My VLA? Demystifying VLA Inference Performance with VLA-Perf},
  author={Jiang, Wenqi and Clemons, Jason and Sankaralingam, Karu and Kozyrakis, Christos},
  journal={arXiv preprint arXiv:2602.18397},
  year={2026}
}

@article{ma2025realtimevla,
  title={Running vlas at real-time speed},
  author={Ma, Yunchao and Zhou, Yizhuang and Yang, Yunhuan and Wang, Tiancai and Fan, Haoqiang},
  journal={arXiv preprint arXiv:2510.26742},
  year={2025}
}

@article{black2025rtc,
  title={Real-time execution of action chunking flow policies},
  author={Black, Kevin and Galliker, Manuel Y and Levine, Sergey},
  journal={arXiv preprint arXiv:2506.07339},
  year={2025}
}

@article{black2025trainrtc,
  title={Training-time action conditioning for efficient real-time chunking},
  author={Black, Kevin and Ren, Allen Z and Equi, Michael and Levine, Sergey},
  journal={arXiv preprint arXiv:2512.05964},
  year={2025}
}

@article{tang2025vlash,
  title={Vlash: Real-time vlas via future-state-aware asynchronous inference},
  author={Tang, Jiaming and Sun, Yufei and Zhao, Yilong and Yang, Shang and Lin, Yujun and Zhang, Zhuoyang and Hou, James and Lu, Yao and Liu, Zhijian and Han, Song},
  journal={arXiv preprint arXiv:2512.01031},
  year={2025}
}

@article{arachchige2025sail,
  title={Sail: Faster-than-demonstration execution of imitation learning policies},
  author={Arachchige, Nadun Ranawaka and Chen, Zhenyang and Jung, Wonsuhk and Shin, Woo Chul and Bansal, Rohan and Barroso, Pierre and He, Yu Hang and Lin, Yingyang Celine and Joffe, Benjamin and Kousik, Shreyas and others},
  journal={arXiv preprint arXiv:2506.11948},
  year={2025}
}

@inproceedings{zitkovich2023rt2,
  title={Rt-2: Vision-language-action models transfer web knowledge to robotic control},
  author={Zitkovich, Brianna and Yu, Tianhe and Xu, Sichun and Xu, Peng and Xiao, Ted and Xia, Fei and Wu, Jialin and Wohlhart, Paul and Welker, Stefan and Wahid, Ayzaan and others},
  booktitle={Conference on Robot Learning},
  pages={2165--2183},
  year={2023},
  organization={PMLR}
}

@article{kim2024openvla,
  title={Openvla: An open-source vision-language-action model},
  author={Kim, Moo Jin and Pertsch, Karl and Karamcheti, Siddharth and Xiao, Ted and Balakrishna, Ashwin and Nair, Suraj and Rafailov, Rafael and Foster, Ethan and Lam, Grace and Sanketi, Pannag and others},
  journal={arXiv preprint arXiv:2406.09246},
  year={2024}
}

@article{black2024pi_0,
  title={{$\\pi_0$}: A Vision-Language-Action Flow Model for General Robot Control},
  author={Black, Kevin and Brown, Noah and Driess, Danny and Esmail, Adnan and Equi, Michael and Finn, Chelsea and Fusai, Niccolo and Groom, Lachy and Hausman, Karol and Ichter, Brian and others},
  journal={arXiv preprint arXiv:2410.24164},
  year={2024}
}

@article{intelligence2025pi_05,
  title={{$\\pi_{0.5}$}: A Vision-Language-Action Model with Open-World Generalization},
  author={Intelligence, Physical and Black, Kevin and Brown, Noah and Darpinian, James and Dhabalia, Karan and Driess, Danny and Esmail, Adnan and Equi, Michael and Finn, Chelsea and Fusai, Niccolo and others},
  journal={arXiv preprint arXiv:2504.16054},
  year={2025}
}

@article{bjorck2025gr00t,
  title={Gr00t n1: An open foundation model for generalist humanoid robots},
  author={Bjorck, Johan and Casta{\~n}eda, Fernando and Cherniadev, Nikita and Da, Xingye and Ding, Runyu and Fan, Linxi and Fang, Yu and Fox, Dieter and Hu, Fengyuan and Huang, Spencer and others},
  journal={arXiv preprint arXiv:2503.14734},
  year={2025}
}

@article{ye2026dreamzero,
  title={World Action Models are Zero-shot Policies},
  author={Ye, Seonghyeon and Ge, Yunhao and Zheng, Kaiyuan and Gao, Shenyuan and Yu, Sihyun and Kurian, George and Indupuru, Suneel and Tan, You Liang and Zhu, Chuning and Xiang, Jiannan and others},
  journal={arXiv preprint arXiv:2602.15922},
  year={2026}
}

@article{kim2026cosmospolicy,
  title={Cosmos policy: Fine-tuning video models for visuomotor control and planning},
  author={Kim, Moo Jin and Gao, Yihuai and Lin, Tsung-Yi and Lin, Yen-Chen and Ge, Yunhao and Lam, Grace and Liang, Percy and Song, Shuran and Liu, Ming-Yu and Finn, Chelsea and others},
  journal={arXiv preprint arXiv:2601.16163},
  year={2026}
}

@inproceedings{guo2025demospeedup,
  title={DemoSpeedup: Accelerating Visuomotor Policies via Entropy-Guided Demonstration Acceleration},
  author={Guo, Lingxiao and Xue, Zhengrong and Xu, Zijing and Xu, Huazhe},
  booktitle={Proceedings of The 9th Conference on Robot Learning},
  series={Proceedings of Machine Learning Research},
  volume={305},
  pages={599--609},
  year={2025},
  publisher={PMLR}
}

@article{hoeg2024streamingdiffusionpolicy,
  title={Streaming Diffusion Policy: Fast Policy Synthesis with Variable Noise Diffusion Models},
  author={H{\o}eg, Sigmund H and Du, Yilun and Egeland, Olav},
  journal={arXiv preprint arXiv:2406.04806},
  year={2024}
}

@article{zhao2023act,
  title={Learning fine-grained bimanual manipulation with low-cost hardware},
  author={Zhao, Tony Z and Kumar, Vikash and Levine, Sergey and Finn, Chelsea},
  journal={arXiv preprint arXiv:2304.13705},
  year={2023}
}

@misc{vial2026drtc, 
    title={Distributed Real-Time Chunking}, 
    url={https://jackvial.com/posts/distributed-real-time-chunking.html}, 
    journal={Distributed Real-Time Chunking - Jack Vial}, 
    author={Vial, Jack},
    year={2026}, 
    month={Mar}
}

@article{zhang2025actionchunking,
  title={Action Chunking and Exploratory Data Collection Yield Exponential Improvements in Behavior Cloning for Continuous Control},
  author={Zhang, Thomas T and Pfrommer, Daniel and Pan, Chaoyi and Matni, Nikolai and Simchowitz, Max},
  journal={arXiv preprint arXiv:2507.09061},
  year={2025}
}

@article{kwok25robomonkey,
  title={RoboMonkey: Scaling Test-Time Sampling and Verification for Vision-Language-Action Models},
  author={Jacky Kwok and Christopher Agia and Rohan Sinha and Matt Foutter and Shulu Li and Ion Stoica and Azalia Mirhoseini and Marco Pavone},
  journal={arXiv preprint arXiv:2506.17811},
  year={2025},
}

@article{liu2023libero,
  title={Libero: Benchmarking knowledge transfer for lifelong robot learning},
  author={Liu, Bo and Zhu, Yifeng and Gao, Chongkai and Feng, Yihao and Liu, Qiang and Zhu, Yuke and Stone, Peter},
  journal={Advances in Neural Information Processing Systems},
  volume={36},
  pages={44776--44791},
  year={2023}
}

@inproceedings{o2024open,
  title={Open x-embodiment: Robotic learning datasets and rt-x models: Open x-embodiment collaboration 0},
  author={O’Neill, Abby and Rehman, Abdul and Maddukuri, Abhiram and Gupta, Abhishek and Padalkar, Abhishek and Lee, Abraham and Pooley, Acorn and Gupta, Agrim and Mandlekar, Ajay and Jain, Ajinkya and others},
  booktitle={2024 IEEE International Conference on Robotics and Automation (ICRA)},
  pages={6892--6903},
  year={2024},
  organization={IEEE}
}

@article{khazatsky2024droid,
  title={Droid: A large-scale in-the-wild robot manipulation dataset},
  author={Khazatsky, Alexander and Pertsch, Karl and Nair, Suraj and Balakrishna, Ashwin and Dasari, Sudeep and Karamcheti, Siddharth and Nasiriany, Soroush and Srirama, Mohan Kumar and Chen, Lawrence Yunliang and Ellis, Kirsty and others},
  journal={arXiv preprint arXiv:2403.12945},
  year={2024}
}

@article{chi2025diffusion,
  title={Diffusion policy: Visuomotor policy learning via action diffusion},
  author={Chi, Cheng and Xu, Zhenjia and Feng, Siyuan and Cousineau, Eric and Du, Yilun and Burchfiel, Benjamin and Tedrake, Russ and Song, Shuran},
  journal={The International Journal of Robotics Research},
  volume={44},
  number={10-11},
  pages={1684--1704},
  year={2025},
  publisher={Sage Publications Sage UK: London, England}
}

@article{dai2026kairos,
  title={Kairos: A Scalable Serving System for Physical AI},
  author={Dai, Yinwei and Ananthanarayanan, Ganesh and Cox, Landon and Foukas, Xenofon and Radunovic, Bozidar and Netravali, Ravi},
  journal={arXiv preprint arXiv:2605.11381},
  year={2026}
}

@article{jiang2026rosa,
  title={ROSA: A Robotics Foundation Model Serving System for Robot Factories},
  author={Jiang, Wenqi and Clemons, Jason and O'Flaherty, Rowland and Hadfield, Hugo and Degirmenci, Alperen and Song, Shuran and Narang, Yashraj and Kozyrakis, Christos},
  journal={arXiv preprint arXiv:2607.01088},
  year={2026}
}

\newpage
\appendix

\section{Appendix}

\localtableofcontents
\newpage

\subsection{Scheduler Pseudocode}
\label{sec:sched_desn}

EDF and RR plan the next batch only when the GPU becomes available. LA, in contrast, searches continuously while the GPU is busy and dispatches its best batch when the GPU becomes available.

\para{Earliest Deadline First.}
EDF greedily fills the batch with the candidates closest to starvation. 

\begin{lstlisting}
def edf(robots: list[Robot]) -> list[Robot]:
  sorted_by_deadline = sorted(robots, key=lambda robot: robot.deadline)
  return sorted_by_deadline[:min(max_batch_size, len(robots))]
\end{lstlisting}

\texttt{robot.deadline} is computed from the mirror as the wall-clock time of the last control step covered by any chunk in $\mathcal{Q}_j$.
 
\para{Round Robin.}
RR cycles through all robots, advancing a pointer across epochs. 
\begin{lstlisting}
i = 0
def round_robin(robots: list[Robot]) -> list[Robot]:
  batch = []
  for _ in range(min(max_batch_size, len(robots))):
      batch.append(robots[i])
      i = (i + 1) % len(robots)
  return batch
\end{lstlisting}

\para{Lookahead.}
LA simulates forward and chooses the first batch from the best scoring schedule.
\begin{lstlisting}
def lookahead(robots: list[Robot]) -> list[Robot]:
    best_schedule, best_score = [], 0
    # (schedule, simulated state)
    frontier = deque([([], robots)])  

    while frontier and has_slack():
        schedule, state = frontier.popleft()
        for batch in candidate_batches(state):
            next_state = simulate(state, batch)
            new_schedule = schedule + [batch]
            score = score_schedule(robots, next_state, new_schedule) 
            if score > best_score:
                best_schedule, best_score = new_schedule, score
            if len(new_schedule) < max_depth:
                frontier.append((new_schedule, next_state))

    return best_schedule[0]
\end{lstlisting}
\texttt{has\_slack} is true while the current time is more than a few milliseconds before the GPU's next anticipated dispatch time; \texttt{candidate\_batches(s)} enumerates batches of size $1, \dots, B_{\max}$; 
\texttt{simulate(s, b)} uses the mirror to roll
the MDP transition of \S\ref{subsec:mdp} to the next epoch; and \texttt{score\_schedule} calculates $\text{Score}(S)$ using the mirror. 

In our implementation, \texttt{candidate\_batches(s)} heuristically prunes batches to reduce the search space and \texttt{score\_schedule} calculates the reward as the total executed robot time inside a 1-second window starting from the schedule's starting time.

\subsection{Other GPUs}
\label{sec:gpu_compare}

We cloud-serve all of our experiments on an L40S datacenter GPU, but it is possible to serve on all other GPU types as well. In this section, we do some preliminary analysis on inference latency between an L40S and an H100. Table \ref{tab:inference_profile_gpu} details the inference latencies as a function of batch size on both GPUs with the $\pi_{0.5}$ model. We note that because $\pi_{0.5}$ leans more compute-bound than memory-bound, an H100 is able to considerably reduce inference latency and allow for smaller deltas between consecutive batch sizes. In practice, the schedulers will be able to schedule larger batches at lower inference cost, supporting larger numbers of robots. 

\begin{table}[h]
\centering
\caption{$\pi_{0.5}$ inference latency (ms) as a function of batch size on L40S and H100 GPUs.}
\label{tab:inference_profile_gpu}
\begin{tabular}{ccc}
\toprule
Batch size & L40S (ms) & H100 (ms) \\
\midrule
1 & 73.0 & 42.3 \\
2 & 109.8 & 56.0 \\
3 & 142.4 & 66.1 \\
4 & 177.5 & 79.3 \\
5 & 211.1 & 87.2 \\
\bottomrule
\end{tabular}
\end{table}

\subsection{Additional robot foundation models}
\label{sec:vla_compare}

\ourengine{} is not limited to $\pi_{0.5}$ as the only model choice; it is possible to configure backends for other popular VLA models, such as NVIDIA's GR00T-N1 \cite{bjorck2025gr00t}. In this section, we do some preliminary analysis on the inference profile of GR00T. Table \ref{tab:inference_profile_pi05} details the inference profiles as a function of batch size. We see that GR00T is more memory-bound than $\pi_{0.5}$, allowing for smaller deltas between inference times for consecutive batch sizes. Similar to the insight from Section \ref{sec:gpu_compare}, the schedulers will be able to support larger numbers of robots by scheduling larger batches at lower inference cost.

\begin{table}[h]
\centering
\caption{Inference latency (ms) as a function of batch size on $\pi_{0.5}$ and GR00T N1.7 on an L40S GPU.}
\label{tab:inference_profile_pi05}
\begin{tabular}{ccc}
\toprule
Batch size & $\pi_{0.5}$ & GR00T N1.7 \\
\midrule
1 & 73.0 & 71.8 \\
2 & 109.8 & 76.5 \\
3 & 142.4 & 83.5 \\
4 & 177.5 & 91.8 \\
5 & 211.1 & 100.9 \\
\bottomrule
\end{tabular}
\end{table}

% \subsection{Choosing the right batch size}
% \label{sec:batch_ablate}

% In our work, we posit that naively using the maximum batch size is not always the best decision, and that many scenarios require dynamic batching. \ours{} is capable of taking a maximum batch size cap and choosing the best batch of robots within that cap to serve, while more naive heuristic schedulers like EDF and Round-Robin batch as many robots together as they can at each decision step. Because of these properties, we ablate over the maximum batch size cap so as to give the naive methods the best shot at efficient scheduling. We found that with our setup of 10 robots, $b=3$ is the best parameter.

% \begin{figure}[h]
%     \centering
%     \includegraphics[width=\linewidth]{figures/batch_size_choice.pdf}
%     \vspace*{1mm}
%     \caption{How does the max batch size affect scheduler performance? We ablate this on the 10-fast scenario, showcasing that \ours{} performs most effectively by choosing smaller/more efficient batch sizes more often.}
%     \label{fig:batch_size_ablation}
    
% \end{figure}

  \subsection{Choosing the right batch size}
  \label{sec:batch_ablate}

\begin{figure}[h]
      \centering
      \includegraphics[width=\linewidth]{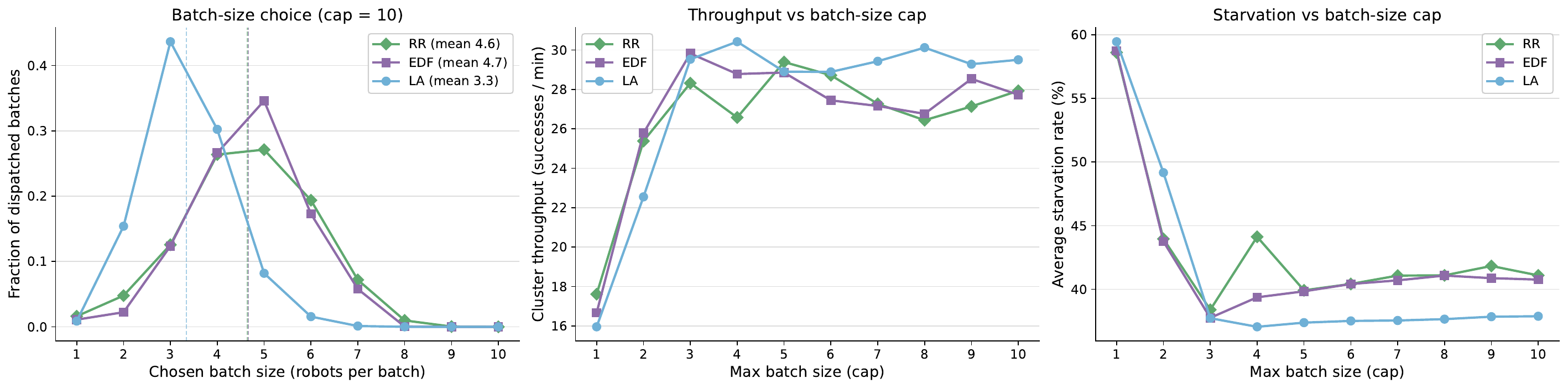}
      \vspace*{1mm}
      \caption{How does the max batch size affect scheduler performance? We ablate
      this on the all-fast scenario. \ours{} performs most effectively by choosing
      smaller, more efficient batches, keeping starvation lower as the batch-size cap grows.}
      \label{fig:batch_size_ablation}
  \end{figure}
  
In our work, we posit that naively using the maximum batch size is not always the best decision, and that many scenarios require dynamic batching. \ours{} can take a maximum batch size cap and choose the best batch of robots within that cap to serve, while more naive heuristic schedulers like EDF and Round-Robin batch as many robots together as they can at each decision step.

The reason this matters is that batching is a tradeoff. Larger batches amortize inference across more robots, but inference latency grows with batch size (Table~\ref{tab:inference_profile_pi05}): on an L40S, a batch of $5$ takes $211$ms versus $73$ms for a single robot, while a batch of $3$ takes only $142$ms and already captures most of the per-robot amortization. A scheduler that always fills the batch therefore serves more robots per call but makes every call slower and less frequent, so robots left out of the current batch wait longer and starve. Choosing the right batch size means balancing this amortization against the latency penalty.
  
We ablate over the maximum batch size cap so as to give the naive methods the best shot at efficient scheduling. We run this study on the all-fast scenario because homogeneous deadlines are the most batching-friendly setting: it is where fixed max-batching is least penalized, making it the strongest case for the naive heuristics.

Figure~\ref{fig:batch_size_ablation} shows the result. The left panel reports the distribution of chosen batch sizes at a cap of $10$: EDF and Round-Robin gravitate to large batches (mean $4.7$ robots per call), whereas \ours{} prefers smaller, more efficient batches (mean $3.3$, median $3$). The center and right panels show that at small caps, all schedulers are forced into small batches and behave similarly; as the cap grows, the naive schedulers exploit it by over-batching, and their starvation rate climbs. \ours{} is largely insensitive to the cap (because it already selects efficient batches, raising the cap does not tempt it into over-batching), holding starvation and throughput flat. This amounts to \ours{} delivering lower starvation and higher throughput than the naive methods.

We found that in simulation, with our setup of $10$ robots, $b=3$ is the best parameter that gives the naive schedulers their strongest configuration while keeping per-call latency low enough to avoid widespread starvation, and we use it as the batch-size cap in all main experiments. In the real world, we found $b=3$ to be best for all-fast, and $b=5$ to be best for one-fast and half-fast.

\subsection{Lookahead $L$}
\label{sec:lookahead_ablate}
For Lookahead, the $L$ parameter determines the length of the schedules (in epochs) the scheduler considers in its search. We originally hypothesized that planning for longer would improve scheduler performance, but we found that this is not the case for our implementation. In Table \ref{tab:depth-robustness}, we ablate $L=1,2,3$ across our experimental setups in LIBERO and find that increasing $L$ slightly increases average starvation. We note that these results are specific to our implementation, and that changes to the search algorithm or reward function may show different results.

\begin{table}[h]
\centering
\caption{Average starvation (\%) slightly increases across lookahead depths $L\in\{1,2,3\}$.}
\label{tab:depth-robustness}
\footnotesize
\setlength{\tabcolsep}{4pt}
\begin{tblr}{
  colspec = {l l *{9}{c}},
  row{1,2} = {font=\bfseries},
  column{1,2} = {font=\bfseries},
  colsep = 4pt,
  rowsep = 1.5pt,
}
\toprule
 &  & \SetCell[c=3]{c} LA@1 &  &  & \SetCell[c=3]{c} LA@3 &  &  & \SetCell[c=3]{c} LA@5 &  &  \\
\cmidrule[lr]{3-5} \cmidrule[lr]{6-8} \cmidrule[lr]{9-11}
Config & $N$ & $L{=}1$ & $L{=}2$ & $L{=}3$ & $L{=}1$ & $L{=}2$ & $L{=}3$ & $L{=}1$ & $L{=}2$ & $L{=}3$ \\
\midrule
\SetCell[r=5]{} All Fast & 2 & 1.0 & 1.1 & 1.0 & 1.0 & 1.2 & 1.0 & 1.0 & 1.0 & 1.0 \\
 & 4 & 11.0 & 11.1 & 10.7 & 10.9 & 11.2 & 11.0 & 10.9 & 10.8 & 10.7 \\
 & 6 & 22.5 & 22.9 & 23.1 & 22.9 & 22.7 & 23.1 & 22.9 & 22.7 & 23.1 \\
 & 8 & 30.4 & 30.7 & 31.3 & 30.6 & 30.7 & 31.4 & 30.5 & 30.9 & 31.3 \\
 & 10 & 36.6 & 37.3 & 38.1 & 36.7 & 37.2 & 38.3 & 36.6 & 37.1 & 37.8 \\
\midrule
\SetCell[r=5]{} One Fast & 2 & 1.0 & 1.0 & 1.1 & 1.0 & 1.0 & 1.0 & 1.1 & 0.9 & 1.1 \\
 & 4 & 4.5 & 4.3 & 4.3 & 3.1 & 3.2 & 3.2 & 3.0 & 3.1 & 3.0 \\
 & 6 & 6.4 & 6.3 & 6.4 & 6.3 & 6.8 & 7.3 & 9.0 & 9.3 & 10.0 \\
 & 8 & 11.6 & 12.0 & 12.3 & 12.5 & 12.8 & 15.0 & 14.5 & 15.6 & 17.7 \\
 & 10 & 17.6 & 18.1 & 18.4 & 18.5 & 19.1 & 21.0 & 21.4 & 22.7 & 24.2 \\
\midrule
\SetCell[r=5]{} Half Fast & 2 & 1.0 & 1.0 & 1.1 & 1.1 & 1.0 & 1.1 & 1.0 & 1.1 & 1.0 \\
 & 4 & 6.8 & 7.0 & 6.7 & 6.7 & 7.2 & 6.8 & 19.5 & 17.2 & 17.3 \\
 & 6 & 13.7 & 13.6 & 13.9 & 14.9 & 15.5 & 16.6 & 22.2 & 22.7 & 23.8 \\
 & 8 & 20.4 & 20.9 & 21.1 & 22.4 & 23.7 & 24.7 & 33.5 & 35.6 & 36.9 \\
 & 10 & 27.2 & 27.4 & 27.6 & 29.1 & 31.3 & 31.4 & 47.4 & 48.9 & 48.9 \\
\bottomrule
\end{tblr}
\end{table}

% \newpage
\subsection{Effect of network on cloud serving}
\label{sec:network_ablate}

A natural concern with cloud-based robot policy serving is whether network latency and jitter make the approach impractical. We ablate both factors in simulation with the all-fast, one-fast, half-fast configurations with $b=3$.

\para{Modeling network deviation.} We model each one-way link delay (the observation uplink $d_\text{obs}$ and the action downlink $d_\text{action}$) as an independent draw from a log-normal distribution. For a link with target median delay $m$ (ms) and jitter parameter $\sigma$, we sample
  \begin{equation}
  d \sim \mathrm{Lognormal}(\mu,\,\sigma^2), \qquad \mu = \ln m,
  \label{eq:lognormal_delay}
  \end{equation}

so that $\ln d \sim \mathcal{N}(\ln m,\,\sigma^2)$. We parameterize by the \emph{median} rather than the mean because the median is invariant to the jitter: $\mathrm{median}(d) = e^{\mu} = m$ for all $\sigma$, allowing us to sweep jitter while holding the typical latency fixed. A log-normal (rather than Gaussian) captures the heavy right tail characteristic of wide-area networks, where delays are strictly non-negative, most packets arrive near or below the median, and occasional congestion produces spikes.

Increasing $\sigma$ widens this tail asymmetrically while leaving the median fixed. Setting $\sigma = 0$ recovers a deterministic delay of exactly $m$. We draw a fresh sample for every request, so each robot sees new uplink and downlink delays at every inference step.

\begin{figure}[h]
    \centering

    \begin{subfigure}[t]{0.48\linewidth}
        \centering
        \includegraphics[width=\linewidth]{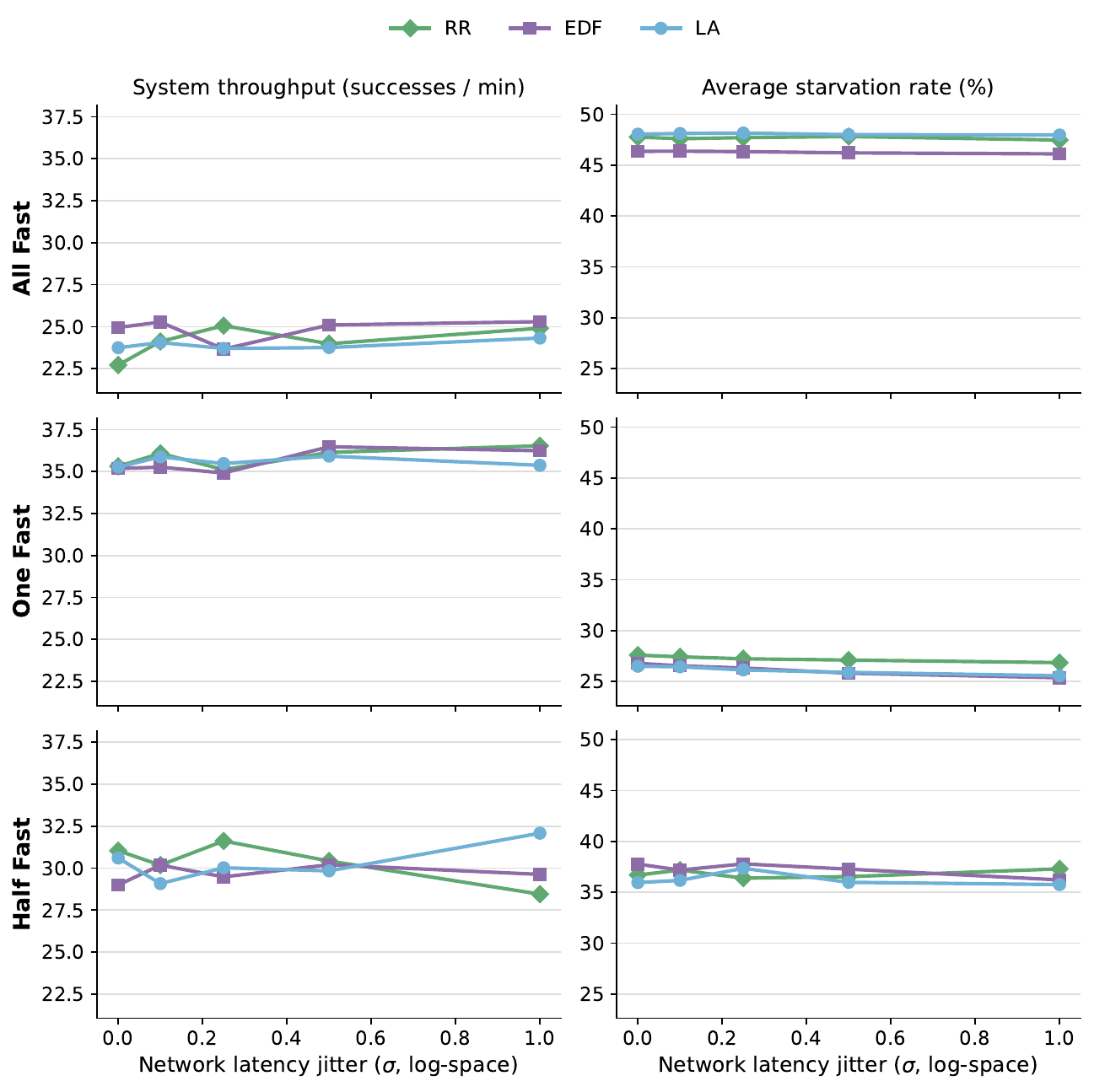}
        \caption{Effect of network jitter variance ($m = 50\text{ms}$).}
        \label{fig:net_ablation_var}
    \end{subfigure}
    \hfill
    \begin{subfigure}[t]{0.48\linewidth}
        \centering
        \includegraphics[width=\linewidth]{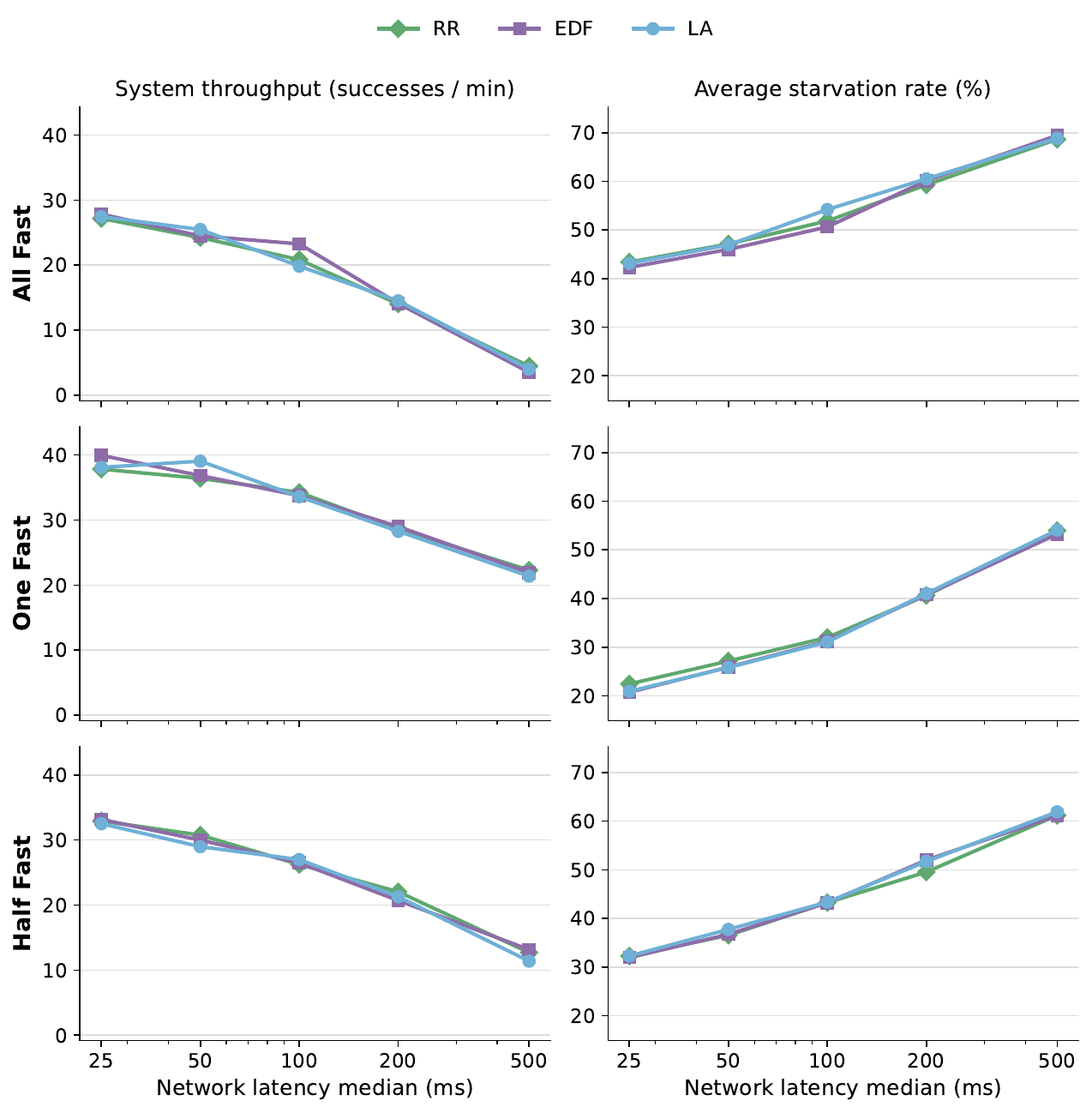}
        \caption{Effect of median network latency ($\sigma=0$).}
        \label{fig:net_ablation_med}
    \end{subfigure}

    \caption{Network ablation study showing how network delay affects cloud serving for $b=3$.}
    \label{fig:network_ablation_jitter}
\end{figure}

\para{Network jitter.} In Figure \ref{fig:net_ablation_var} we study the effect of network jitter on cloud serving. We sweep the standard deviation $\sigma$ of a log-normal jitter distribution applied to both $d_\text{obs}$ and $d_\text{action}$, while setting each one-way median to \emph{50ms} (standard US coast-to-coast transmission delay). Across the full range of jitter magnitudes, starvation rate and system throughput remain nearly flat for all three schedulers. This stability arises because \ourengine{}'s EMA-based delay estimator absorbs moderate jitter, and the action queue provides a natural buffer against individual delayed packets. The result indicates that our cloud serving system is robust even under high network variance.

\para{Network median delay.} In Figure \ref{fig:net_ablation_med} we study the effect of median network latency on cloud serving, sweeping the median delay applied to both $d_\text{obs}$ and $d_\text{action}$ from 25ms to 500ms with zero jitter. In contrast to the jitter sweep, median latency has a direct and monotonic effect: as latency grows, system throughput decreases and starvation rate increases for all three schedulers and across all robot configurations. The degradation is graceful at the latencies encountered in real deployments. We can see that performance remains high through 50-100ms, spanning typical intra-continental round trips. All three schedulers degrade in the same manner, indicating that generally, scheduling policy cannot compensate for raw transport delay. These results confirm that cloud serving remains practical across the latency regimes of real wide-area deployments, with performance degrading predictably only as latency approaches the extremes.

\subsection{Simulation system metrics}
\label{sec:sim_metrics}

\begin{figure}[h]
    \centering
    \includegraphics[width=\linewidth]{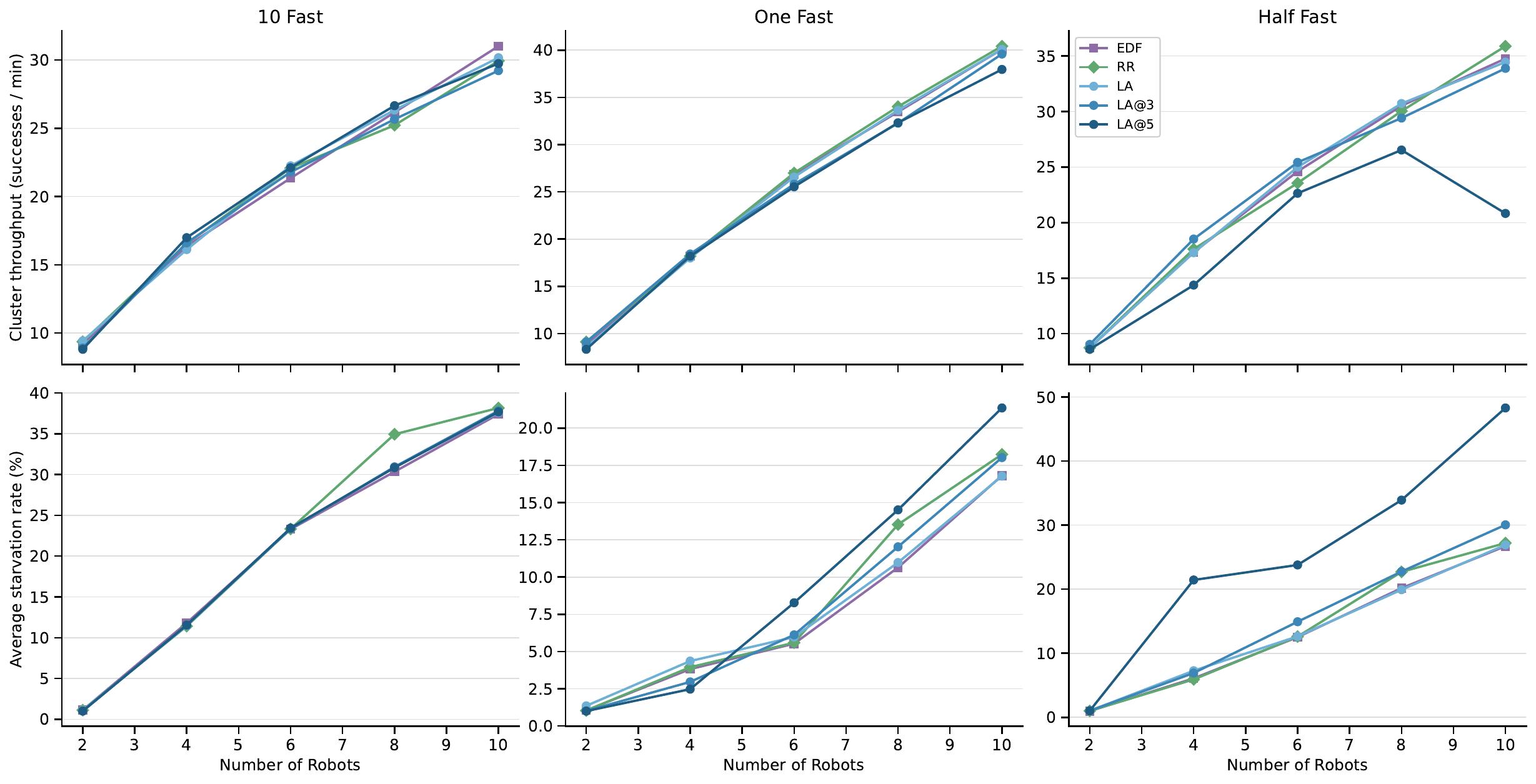}
    \caption{System throughput (successes/min) and average starvation as the cluster scales from $N{=}2$ to $10$ robots, with batch-size cap $b{=}3$.}   
      \label{fig:sim_system_metrics}
\end{figure}

\para{System metrics.} Figure~\ref{fig:sim_system_metrics} and Table~\ref{tab:sys_thr_starv_mbs3} report system throughput and average starvation as we scale the cluster from $N{=}2$ to $10$ robots across the three robot configurations. Throughput grows with $N$ while starvation rises as more robots contend for the shared server, and at $b{=}3$ all schedulers achieve comparable \emph{total} throughput and starvation. The core contribution of \ours{} is not in raising aggregate throughput, but in reallocating it across tiers, giving fast and slow robots a more favorable share of the system's throughput (Section \ref{subsec:evaluations}).

\begin{table}[h]
\centering
\caption{LIBERO: System throughput (successes/min) and average starvation (\%, \textcolor{gray}{gray}), max batch size = 3.}
\label{tab:sys_thr_starv_mbs3}
\footnotesize
\setlength{\tabcolsep}{4pt}
\begin{tblr}{
  colspec = {l l *{5}{c}},
  row{1} = {font=\bfseries},
  column{1,2} = {font=\bfseries},
  colsep = 4pt,
  rowsep = 1.5pt,
}
\toprule
Config & $N$ & RR & EDF & LA $w{=}1$ & LA $w{=}3$ & LA $w{=}5$ \\
\midrule
\SetCell[r=5]{} All Fast & 2 & \cell{9.36}{1.1} & \cell{9.20}{1.1} & \cell{9.42}{1.1} & \cell{8.90}{1.0} & \cell{8.80}{1.0} \\
 & 4 & \cell{16.55}{11.4} & \cell{16.37}{11.8} & \cell{16.11}{11.5} & \cell{16.60}{11.5} & \cell{16.99}{11.6} \\
 & 6 & \cell{22.08}{23.3} & \cell{21.35}{23.3} & \cell{22.25}{23.4} & \cell{21.79}{23.4} & \cell{22.11}{23.4} \\
 & 8 & \cell{25.22}{34.9} & \cell{26.17}{30.3} & \cell{26.34}{30.9} & \cell{25.67}{30.9} & \cell{26.66}{30.8} \\
 & 10 & \cell{29.94}{38.2} & \cell{31.02}{37.4} & \cell{30.18}{37.6} & \cell{29.23}{37.8} & \cell{29.75}{37.7} \\
\midrule
\SetCell[r=5]{} One Fast & 2 & \cell{9.12}{1.0} & \cell{8.83}{1.1} & \cell{8.56}{1.4} & \cell{9.11}{1.0} & \cell{8.33}{1.0} \\
 & 4 & \cell{18.19}{3.9} & \cell{18.19}{3.8} & \cell{18.00}{4.4} & \cell{18.42}{3.0} & \cell{18.19}{2.5} \\
 & 6 & \cell{26.99}{5.6} & \cell{26.87}{5.5} & \cell{26.57}{6.0} & \cell{25.82}{6.1} & \cell{25.54}{8.3} \\
 & 8 & \cell{34.02}{13.5} & \cell{33.47}{10.6} & \cell{33.64}{11.0} & \cell{32.28}{12.0} & \cell{32.31}{14.5} \\
 & 10 & \cell{40.42}{18.2} & \cell{40.10}{16.8} & \cell{40.09}{16.8} & \cell{39.59}{18.0} & \cell{37.96}{21.4} \\
\midrule
\SetCell[r=5]{} Half Fast & 2 & \cell{8.74}{1.0} & \cell{8.73}{1.0} & \cell{8.75}{1.0} & \cell{9.04}{1.0} & \cell{8.60}{1.0} \\
 & 4 & \cell{17.62}{5.9} & \cell{17.33}{6.1} & \cell{17.29}{7.3} & \cell{18.52}{6.9} & \cell{14.38}{21.4} \\
 & 6 & \cell{23.56}{12.6} & \cell{24.61}{12.5} & \cell{25.00}{12.6} & \cell{25.42}{14.9} & \cell{22.64}{23.8} \\
 & 8 & \cell{30.04}{22.7} & \cell{30.52}{20.2} & \cell{30.72}{19.9} & \cell{29.41}{22.8} & \cell{26.54}{33.9} \\
 & 10 & \cell{35.88}{27.2} & \cell{34.76}{26.7} & \cell{34.45}{26.9} & \cell{33.89}{30.0} & \cell{20.83}{48.3} \\
\bottomrule
\end{tblr}
\end{table}

\newpage
\subsection{Real world per-tier metrics}
\label{sec:real_metrics}
In Table \ref{tab:real_all_metrics}, we list the throughputs and starvations from our real-world experiments. In many cases, we find that lower starvation correlates with higher throughputs, matching the results in Figure \ref{fig:throughput_starvation}. However, there are some cases where the Lookahead scheduler has both higher starvations \textit{and} throughputs. This can be clearly seen in the \textbf{All Fast} configuration, where LA's throughput increases by more than 7\% despite a 5.71 percentage-point increase in starvation when compared to EDF.

We suggest a few potential sources for this discrepancy. First, our metrics may be noisy due to small sample size. Setting up tasks for 10 robots in the real-world is very time-consuming, so we are only able to run 1-minute rollouts for 3 seeds.  Second, real-world evaluations are also noisy. While we made our best effort at standardizing task and robot initializations, runs may still differ in their policy predictions as well as the exact scheduling decisions. 

Lastly, our choice of \textit{naive async} as our action chunking strategy may confound with increases in starvation. Prior works have noted that naive async creates discontinuous/jerky motions when execution starts from the middle of a chunk. Intuitively, this is because chunks predicted with naive async have no awareness of actions executed from previous chunks.

\begin{figure}
    \centering
    \includegraphics[width=1\linewidth]{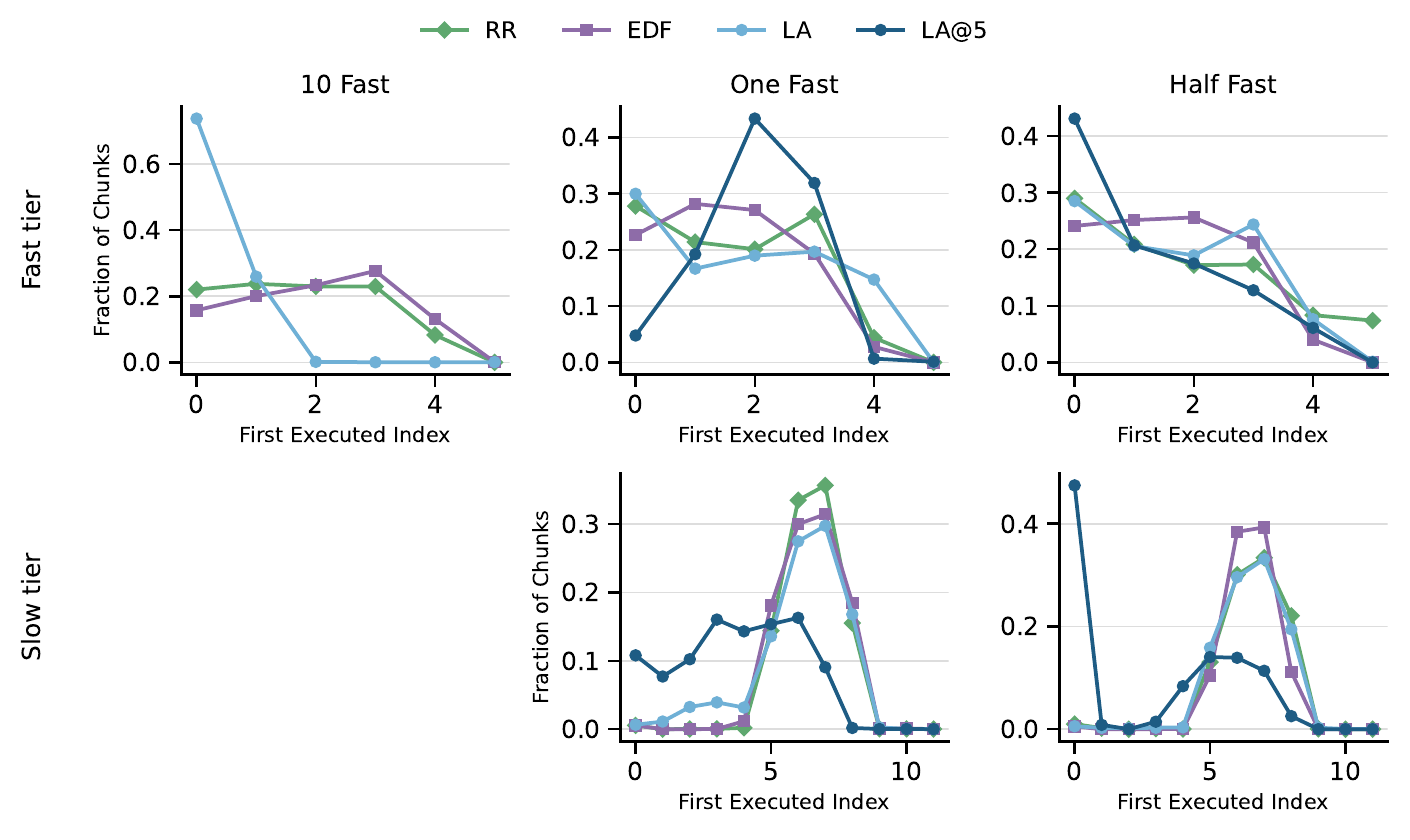}
    \caption{Lookahead schedulers bias towards scheduling chunks so that they are executed starting at index 0.}
    \label{fig:first_executed_index}
\end{figure}

In Figure \ref{fig:first_executed_index}, we plot the distribution of the first executed indexes of chunks across the real-world experiments. The Lookahead schedulers bias towards scheduling a large proportion of chunks to start execution from the start of the chunk (first executed index 0), whereas other schedulers start execution from the middle. We hypothesize that this unintentionally helps robots under Lookahead by reducing the chances for naive async to create jerky motions. Predicting chunks with more sophisticated asynchronous strategies such as RTC~\cite{black2025rtc} or VLASH~\cite{tang2025vlash} may lead to different results.

\begin{table}[h]
\centering
\caption{Real-world: per-tier throughput (successes / min) and starvation rate (\%) across scenarios.}
\label{tab:real_all_metrics}
\footnotesize
\setlength{\tabcolsep}{4pt}
\begin{tblr}{
  colspec = {l l *{6}{c}},
  row{1,2} = {font=\bfseries},
  column{1,2} = {font=\bfseries},
  colsep = 4pt,
  rowsep = 1.5pt,
}
\toprule
 &  & \SetCell[c=3]{c} Throughput (successes/min) &  &  & \SetCell[c=3]{c} Starvation (\%) &  &  \\
\cmidrule[lr]{3-5} \cmidrule[lr]{6-8}
Config & Scheduler & fast & slow & total & avg & fast & slow \\
\midrule
\SetCell[r=3]{} All Fast & EDF & 6.27 & -- & 62.67 & 35.94 & 35.94 & -- \\
 & RR & 6.00 & -- & 60.00 & 44.46 & 44.46 & -- \\
 & LA $w{=}1$ & 6.73 & -- & 67.33 & 41.65 & 41.65 & -- \\
\midrule
\SetCell[r=4]{} One Fast & EDF & 3.67 & 8.81 & 83.00 & 13.37 & 43.27 & 10.05 \\
 & RR & 2.00 & 9.04 & 83.33 & 12.90 & 43.31 & 9.52 \\
 & LA $w{=}1$ & 3.67 & 9.85 & 92.33 & 12.37 & 42.68 & 9.01 \\
 & LA $w{=}5$ & 8.33 & 10.04 & 98.67 & 15.18 & 13.10 & 15.41 \\
\midrule
\SetCell[r=4]{} Half Fast & EDF & 3.13 & 9.40 & 62.67 & 24.93 & 42.09 & 7.78 \\
 & RR & 3.87 & 9.07 & 64.67 & 24.66 & 41.26 & 8.05 \\
 & LA $w{=}1$ & 3.60 & 9.67 & 66.33 & 25.55 & 42.69 & 8.41 \\
 & LA $w{=}5$ & 6.13 & 6.53 & 63.33 & 39.42 & 32.72 & 46.13 \\
\bottomrule
\end{tblr}
\end{table}

\newpage
\subsection{Implementation Details}
\label{sec:implementation}
% \para{Client} Unlike LLMs and other ML serving applications, robots act in real-time and continuously receive feedback from the environment. The robot client communicates with the server over a bidirectional WebSocket connections, sending observations and receiving action chunks. To ensure the server always has the most recent observation to provide to the policy, the robot client continually sends the captured observation to the server when it is captured at every control step. 

Following standard practice in LLM serving engines, \ourengine{} is implemented as a collection of Python processes that communicate over shared memory. 

The \textbf{frontend} process is responsible for communicating with robots over network. Robots initiate a WebSocket connection with the frontend and send the captured observation to the server on each control step. The frontend writes the raw data (images, prompt, metadata) to shared memory and sends the metadata to the scheduler and engine process. 

The \textbf{scheduler} process is responsible for maintaining the software mirror of the robots and scheduling the next batches for the server to execute. The mirror tracks all the robots as well as all the chunks. For EDF and RR, the scheduler schedules the next batch when the GPU becomes available, while the Lookahead scheduler continually searches and queues the best schedule whenever the GPU becomes available again.

The \textbf{engine} process is a worker that is solely responsible for running model forward passes. On startup, it profiles $d_{\text{infer}}$ for each batch size and sends it to the scheduler. Afterwards, it busy-waits until a batch arrives from the scheduler. For each batch it consumes, it directly reads the latest request in shared memory. If the latest request is identical to the last served request, or the robot has not executed $H_\text{min}$ actions since the last served request, the request will be dropped from the batch. For the $\pi_{0.5}$ forward pass, \ourengine{} uses the official JAX implementation~\cite{intelligence2025pi_05}.

\end{document}